\documentclass[10pt,twocolumn,letterpaper]{article}

\usepackage[pagenumbers]{cvpr} 

\usepackage[utf8]{inputenc} 
\usepackage[T1]{fontenc}    
\usepackage{url}            
\usepackage{booktabs}       
\usepackage{amsfonts}       
\usepackage{nicefrac}       
\usepackage{microtype}      
\usepackage{xcolor}         

\usepackage{enumitem}
\usepackage{booktabs}
\usepackage{multirow} 
\usepackage{enumitem}
\usepackage{colortbl} 
\usepackage{duckuments} 
\usepackage{xspace}
\usepackage{nicefrac}

\usepackage{algorithm}
\usepackage{algpseudocode}                  
\usepackage{setspace}                       
\usepackage{lipsum} 

\usepackage{mathtools} 
\usepackage{bm} 
\usepackage{soul} 
\usepackage{wrapfig}

\usepackage[pagebackref,breaklinks,colorlinks,allcolors=cvprblue]{hyperref}

\newcommand{\bc}{\mathbf{c}}

\newcommand{\bx}{\mathbf{x}}
\newcommand{\bz}{\mathbf{z}}

\newcommand{\bh}{\mathbf{h}}

\newcommand{\lname}{Memory-Augmented Latent Transformers\xspace}
\newcommand{\sname}{MALT\xspace}

\newcommand*{\ShowNotes}{} 

\ifdefined\ShowNotes
  \newcommand{\colornote}[3]{{\color{#1}\bf{#2: #3}\normalfont}}
\else
  \newcommand{\colornote}[3]{}
\fi

\def\paperID{2384} 
\def\confName{CVPR}
\def\confYear{2025}

\title{MALT Diffusion: Memory-Augmented Latent Transformers \\
for Any-Length Video Generation}

\author{%
  {Sihyun Yu$^{1}$\thanks{Work done at Google Research.} \quad Meera Hahn$^2$  \quad Dan Kondratyuk$^{4\,\ast}$ \quad Jinwoo Shin$^1$} \\
  {Agrim Gupta$^{2}$ \quad Jos\'e Lezama$^2$ \quad Irfan Essa$^{2,3}$
  \quad David Ross$^{2}$ \quad Jonathan Huang$^{5\,\ast}$} \\ 
  {$^1$KAIST \quad $^2$Google DeepMind \quad $^3$Georgia Tech \quad $^4$Luma AI \quad $^5$General Robotics}
}

\definecolor{cornellred}{rgb}{0.7, 0.11, 0.11}
\definecolor{cadmiumgreen}{rgb}{0.0, 0.42, 0.24}

\hypersetup{
  linkcolor = cornellred,
  citecolor  = cadmiumgreen,
  colorlinks = true,
}

\begin{document}
\maketitle
\begin{abstract}
Diffusion models are successful for synthesizing high quality videos but are limited to generating short clips (\eg~2-10 seconds).
Synthesizing sustained footage (\eg~over minutes) still remains an open research question.  
In this paper, we propose \emph{\sname Diffusion} (using \lname), a new diffusion model specialized for long video generation. 
\sname Diffusion (or just \sname) handles long videos by subdividing them into short segments and doing segment-level autoregressive generation.
To achieve this, we first propose recurrent attention layers that encode multiple segments into a compact memory latent vector; by maintaining this memory vector over time, \sname is able to condition on it and continuously generate new footage based on a long temporal context.
We also present several training techniques that enable the model to generate frames over a long horizon with consistent quality and minimal degradation.
We validate the effectiveness of \sname through experiments on long video benchmarks.
We first perform extensive analysis of \sname in long-contextual understanding capability and stability using popular long video benchmarks.
For example, \sname achieves an FVD score of 220.4 on 128-frame video generation on UCF-101, outperforming the previous state-of-the-art of 648.4.
Finally, we explore \sname’s capabilities in a text-to-video generation setting and show that it can produce long videos compared with recent techniques for long text-to-video generation.
\end{abstract}
\vspace{-0.1in}
\section{Introduction}
\label{sec:intro}
Diffusion models (DMs)~\citep{ho2021denoising,song2021scorebased} provide a scalable approach to training high quality generative models and have been shown in recent years to apply widely across a variety of data domains including images~\citep{dhariwal2021diffusion,karras2022edm}, audio~\citep{kong2020diffwave,lakhotia2021generative}, 3D shapes~\citep{lee2024dreamflow, luo2021diffusion, zeng2022lion}, and more. Inspired by these successes, a number of DM-based approaches have also been proposed for generating videos~\citep{blattmann2023align,gupta2023photorealistic,ho2022video,bar2024lumiere} and even enabling complex zero-shot text-to-video generation \citep{gupta2023photorealistic,videoworldsimulators2024}.

\begin{figure*}[t!]
    \centering    
    \vspace{-0.05in}
    \includegraphics[width=\linewidth]{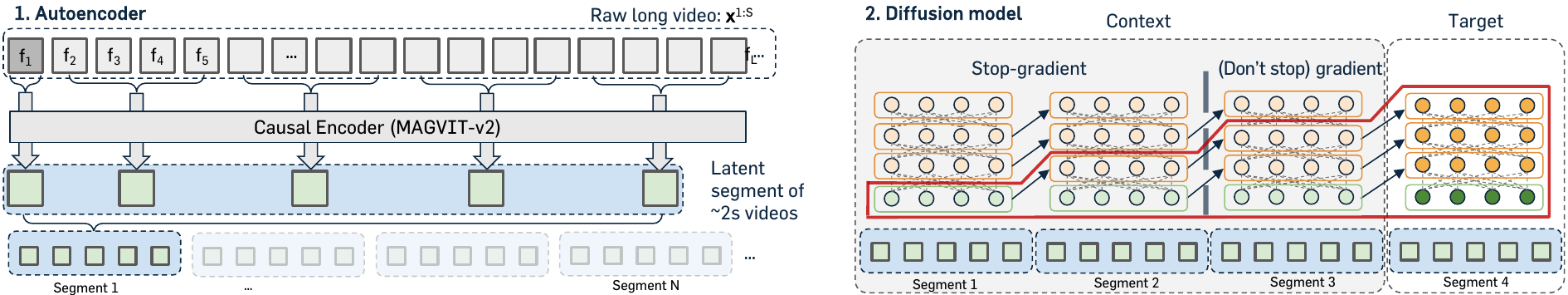}
    \vspace{-0.1in}
    \caption{\textbf{Overview.} \sname first encodes a long video into a low-dimensional latent vector using a video autoencoder
    (\eg, MAGVIT-v2~\citep{yu2024language}) and then divides it into
    multiple segments. After that, we recurrently generate these segments
    one-by-one, based on our newly proposed diffusion transformer
    architecture by adding recurrent temporal attention
    layers to existing architectures~\citep{gupta2023photorealistic,Peebles2022DiT}.
    The red line on right figure denotes the increased receptive field of the \sname relative to a single segment.
    }
    \label{fig:concept}
\end{figure*}

In video generation, a central challenge and key desiderata is to achieve \emph{any-length} generation where the model is capable of producing a video of arbitrary length conditioned on its understanding of the previous context. Many state-of-the-art existing video DM frameworks are \emph{prima facie}, only capable of generating videos of fixed temporal length (\eg, in the form of a fixed shape 3D RGB tensor \citep{ho2022video}) and generally limited from 2-10 seconds. A standard ``naive’’ approach for addressing any-length video generation is to use autoregressive generation where new frames are generated conditioned on a small number of previously generated video frames (rather than a long context). For example, given a 4-second base model, one might autoregressively generate 2 new seconds of footage at a time, conditioning on the previous 2 seconds. Thus, this autoregressive approach can generate compelling videos for longer extents than the base model (especially for very high-quality generation models). However, it also has clear limitations. By not conditioning on long temporal extents, these models often struggle to achieve long-term consistency.  Moreover, small errors during generation can easily cause error propagation~\citep{ruhe2024rolling}, resulting in a rapid quality drop with the frame index.

To summarize, there are two key subchallenges entailed by the any-length video generation problem:
\begin{itemize}[leftmargin=*,itemsep=0mm]
    \item \textbf{long-term contextual understanding}: allowing the model to condition on long temporal extents while staying within memory limitations and computational constraints
    \item \textbf{long-term stability}: generating long videos without frame quality degradation over time.
\end{itemize}
Solving both of these issues is a critical next step for video generation, and indeed, there have been a number of recent works that generate stable video using long context~\citep{harvey2022flexible,yan2023temporally}. However, they have been limited to simple datasets and fail to be stable for complex, realistic datasets.

\vspace{0.02in}
\noindent\textbf{Our approach.}
In this paper, we tackle any-length video generation by achieving the aforementioned twin goals of long-term contextual understanding and long-term stability. Toward this end, we introduce a new latent DM specialized for long videos, coined \emph{\sname Diffusion}: \lname (\sname for short); see Figure~\ref{fig:concept} for an illustration. For long-term contextual understanding, we introduce a new diffusion transformer architecture that enables blockwise autoregressive generation of long videos. Motivated by recent language models developed for handling long sequences~(\eg, \citep{dai2019transformer}), we design our model by adding recurrent temporal attention layers into an existing diffusion transformer architecture, which computes the attention between the current segment and the hidden states of previous segments (context). Hence, hidden states of contextual segments act as a ``memory'' to encode context and is used to generate future segments. Consequently, in contrast to conventional DMs, our model can behave as both a DM for generating a video clip and as an encoder to encode previous contexts (clean inputs) as memory.

To mitigate error accumulation at inference (\ie, to achieve long-term stability), we propose a training technique that
encourages our trained model to be robust with 
respect to noisy memory vectors by applying noise augmentation to the memory vector at training time. Moreover, we carefully design our model optimization to mitigate the rapid increase of
computation and memory costs with respect to video length, enabling the training to be done on very long videos. 
With these components, \sname outperforms previous state-of-the-art results on popular long video generation and prediction benchmarks \citep{yu2023video,yan2023temporally} while using about $2\times$ fewer parameters than the prior best baselines.\footnote{We mainly focus on videos, but ideas can be similarly applied to other domain like PDE or climate data~\citep{ruhe2024rolling}.}

\vspace{0.02in}
\noindent{The main contributions of our paper are:}

\begin{itemize}[leftmargin=*,itemsep=0mm]
    \item We propose a novel latent diffusion model \emph{\sname Diffusion}: \lname (or just \sname), which can generate and be trained on long videos and addresses the aforementioned shortcomings. 
    \item We propose several techniques for training our architecture on long videos, resulting in a robust model both capable of being a denoiser and an encoder of previous contexts. We also ensure training can be performed without expensive memory requirements.
    \item We show the strengths of \sname 
    on long video generation/prediction benchmarks. For instance, \sname achieves an FVD~\citep{unterthiner2018towards} of 220.4 in unconditional 128-frame video generation on UCF-101~\citep{soomro2012ucf101}, 66.0\% better than the previous state-of-the-art of 648.4. Also, on a challenging realistic dataset, Kinetics-600~\citep{kay2017kinetics}, \sname exceeds the prior best long video prediction FVD as 799$\to$392. 
    \item We also qualitatively validate \sname's capability in long-term contextual understanding and stability.
    In particular, we show that \sname shows a strong length generalization: \sname generates long videos when trained on an open-world short text-to-video dataset (up to 4-5 seconds 
    per clip), producing >120 seconds at 8 fps without suffering from significant frame quality degradation.
\end{itemize}

\section{\sname Diffusion with \lname}
\label{sec:method}
Consider a dataset $\mathcal{D}$, where each example $(\bx, \bc) \in \mathcal{D}$ consists of a video $\bx$ and corresponding conditions $\bc$ (\eg, text captions). Our goal is to train a model using $\mathcal{D}$ to learn a model distribution $p_{\text{model}}(\bx | \bc)$ that matches a ground-truth conditional distribution $p_{\text{data}} (\bx | \bc)$. In particular, we are interested in the situation where each $\bx$ is a \emph{long} video, and video lengths are much larger than conventional methods that often use $\sim$20-128 frames for both training and inference~\citep{he2022lvdm}.

To efficiently model long video distribution, we adopt a ``memory'' that encodes previous long context in our latent diffusion transformer. Specifically, we aim to train a single model capable of: (a) encoding previous context of the long video as a compact memory latent vector, and (b) generating a future clip conditioned on the memory and $\bc$. 

In the rest of this section, we explain \sname Diffusion in detail. In Section~\ref{subsec:ldm}, we provide a brief overview of latent diffusion models. In Section~\ref{subsec:obj}, we describe problem formulation and how we design a training objective. Finally, in Section~\ref{subsec:arch}, we explain the architecture that we used.

\vspace{0.02in}
\noindent\textbf{Notation.}
We write a sequence of vectors $[\bx^{a} \ldots, \bx^{b}]$ with $a<b$ as $\bx^{a:b}$. 

\subsection{Latent diffusion models}
\label{subsec:ldm}
In order to generate data, 
diffusion models learn the \emph{reverse} process of a forward diffusion, where the forward diffusion diffuses an 
example $\bx_{0} \sim p_{\text{data}}(\bx)$ to a (simple) prior distribution $\bx_{T} \sim \mathcal{N}(\mathbf{0}, \sigma_{\mathrm{max}}^2 \mathbf{I})$ (with pre-defined $\sigma_{\mathrm{max}}>0$) with the following stochastic differential equation (SDE):
\begin{align}
    d\bx = \mathbf{f} (\bx, t) dt + g(t) d\mathbf{w},
    \label{eq:forwardsde}
\end{align}
where $\mathbf{f}$, $g$, $\mathbf{w}$ are pre-defined drift and diffusion coefficients, and standard Wiener process (respectively) with $t \in [0, T]$ with $T>0$. 
With this forward process, data sampling can be done with the following reverse SDE of Eq.~\eqref{eq:forwardsde}: 
\begin{align}
    d\bx = \Big[ \mathbf{f} (\bx, t) - \frac{1}{2} g(t)^2 \nabla_{\bx} \log p_t (\bx) \Big] dt + g(t) d\mathbf{\bar{w}},
\end{align}
where $\mathbf{\bar{w}}$ is a standard reverse-time Wiener process, and $\nabla_{\bx} \log p_t(\bx)$ is a score function of the marginal density from Eq.~\eqref{eq:forwardsde} at time $t$.
\citet{song2021scorebased} shows there exists a \emph{probability flow ordinary differential equation (PF ODE)}
whose marginal $p_t (\bx)$ is identical for the SDE:
\begin{align}
    d\bx = \Big[ \mathbf{f} (\bx, t) - \frac{1}{2} g(t)^2 \nabla_{\bx} \log p_t (\bx) \Big] dt.
    \label{eq:pfode}
\end{align}
Following previous diffusion model methods~\citep{lee2024dreamflow,zheng2024fast}, we use the setup in EDM~\citep{karras2022edm} with $\mathbf{f} (\bx, t) \coloneqq \mathbf{0}$, $g(t)\coloneqq \sqrt{2\dot\sigma(t)\sigma(t)}$ and a decreasing noise schedule $\sigma: [0, T] \to \mathbb{R}_{+}$. In this case, the PF ODE in Eq.~\eqref{eq:pfode} can be written:
\begin{align}
    d\bx = -\dot\sigma(t)\sigma(t)\nabla_{\bx}\log p (\bx; \sigma(t)) dt,
\end{align}
where  $p(\bx; \sigma)$ is
the smoothed distribution by adding i.i.d Gaussian noise $\bm{\epsilon}\sim\mathcal{N}(\mathbf{0}, \sigma^2 \mathbf{I})$ with standard deviation $\sigma > 0$. To learn the score function $\nabla_{\bx}\log p (\bx; \sigma(t))$, we train a denoising network $D_{\bm{\theta}} (\bx, t)$ with the denoising score matching (DSM)~\citep{song2019generative} objective for all $t \in [0, T]$:
\begin{align*}
\mathbb{E}_{\bx, \bm{\epsilon}, t}\Big[ \lambda({t}) || D_{\bm{\theta}}(\bx_0 + \bm{\epsilon}, t)  - \mathbf{\bx}_0||_2^2 \Big],
\end{align*}
where $\lambda(\cdot)$ assigns a non-negative weight.

However, training $D_{\bm{\theta}}$ directly with raw high-dimensional $\bx$ is computation and memory expensive. To solve this problem, latent diffusion models~\citep{rombach2021highresolution} first learn a 
lower dimensional latent representation of $\bx$ by training an autoencoder (with encoder $F(\bx) = \bz$ and 
decoder $G(\bz) = \bx)$ to reconstruct $\bx$ from 
the low-dimensional vector $\bz$, and then train $D_{\bm{\theta}}$ to generate in this latent space instead.
Specifically, latent diffusion models use the following denoising objective defined in the latent space:
\begin{align*}
    \mathbb{E}_{\bx, \bm{\epsilon}, t}\Big[ \lambda(t) || D_{\bm{\theta}}(\bz_0 + \bm{\epsilon}, t)  - \mathbf{\bz}_0||_2^2 \Big],
\end{align*}
where $\bz_0=F(\bx_0)$. After training the model in latent space, we sample a latent vector $\bz$ through an ODE or SDE solver~\citep{song2021denoising,song2021scorebased,karras2022edm} starting from random noises and then decode the result using $G$ to generate
a final sample.

\subsection{Modeling long videos via blockwise diffusion}
\label{subsec:obj}

Given a ``long'' video $\bx^{1:S} \in \mathbb{R}^{S \times H \times W \times 3}$ with a resolution $H \times W$ and number of frames $S>0$, we divide the video into $N$ \emph{segments} of length $L$: $\bx^{(i-1)L+1:iL}$ for $1\leq i \leq N$ (thus $S=NL$). \sname will autoregressively generate
these segments one-by-one.

\vspace{0.02in}
\noindent\textbf{Autoencoder.}
Following the standard latent diffusion approach, we would like to be able to encode and decode 
videos using a trained autoencoder, which is typically more lightweight 
compared to the diffusion model.  However for very long videos, even running the autoencoder
can be infeasible---thus we encode and decode videos in \emph{chunks} of $m<N$ contiguous segments at a time.  Specifically, for $1 \leq i \leq N/m$, we encode $\bx^{(i-1)mL+1:imL}$ as a latent vector $\bz^{im:(i+1)m}$ and decode it as:
\begin{gather*}
    \bz^{im:(i+1)m}\coloneqq F(\bx^{(i-1)mL+1:imL}) \in \mathbb{R}^{m\cdot l \times h \times w \times c}, \\
    G(\bz^{im:(i+1)m})\approx \bx^{(i-1)mL+1:imL},
\end{gather*}
where $F(\cdot)$ is an encoder network that maps the original video segments to their corresponding latent vectors with a spatial downsampling factor $d_s = H/h = W/w > 1$ and a temporal downsampling factor $d_l = L/l > 1$, and $G(\cdot)$ is a decoder network. We use these latent segments $\bz^{1}, \ldots, \bz^{n}$ of the original $\bx$ obtained from the autoencoder for modeling the long video distribution.

\vspace{0.02in}
\noindent\textbf{Diffusion model.}
We now directly model the joint distribution of latent segments,  $p(\bz^{1:N}|\bc)$, autoregressively as 
\begin{align}
p (\bz^{1:N} | \bc) = \prod_{n=0}^{N-1} p(\bz^{n+1} | \bz^{1:n}, \bc) \text{with}\,\,\, \bz^{1:0}\coloneqq \mathbf{0}
\end{align}
where we learn all 
$p(\bz^{n+1} | \bz^{1:n}, \bc)$ for $0\leq n \leq N-1$ using a single diffusion model $D_{\bm{\theta}}$.

A na\"ive approach would be to use $\bz^{1:n}$ directly as a condition to the model; however, if the number of segments, $N$, is large, $\bz^{1:n}$ easily becomes extremely high-dimensional and prohibitive with respect to memory and compute. To mitigate this bottleneck, we instead introduce a \emph {fixed-size} hidden state $\bh^{i} \coloneqq [\bh^{i}_1, \ldots, \bh^{i}_{d}]$ recurrently computed from $D_{\bm{\theta}}$. This hidden state serves as a memory vector to encode the context (i.e., the previous sequence of segments) $\bz^{1:n}$, where $d>0$ is a number of hidden states that are used as memory vectors (\ie, $i$ is a segment index and $d$ refers to the number of layers of the model). 

Specifically, for $1 \leq i \leq n$, we compute $\bh^n$ with the following recurrent mechanism:
\begin{equation}
    \bh^{i} = \mathrm{HiddenState}\big(D_{\bm{\theta}} (\bz_0^{i}, 0;\, \small{\mathtt{sg}}(\bh^{i-1}), \bc)\big),
\end{equation}
where $\bh^{0} = [\mathbf{0}, \ldots, \mathbf{0}]$ and $\small{\mathtt{sg}}$ denotes a stop-grad operation.
Note that we use a clean segment $\bz^{i}$ without added noise so we set $t=0$ here, which has not been used in conventional diffusion model training~\citep{ho2021denoising,song2021denoising,song2021scorebased}.

\begin{figure}[t!]
    \centering    
    \includegraphics[width=\linewidth]{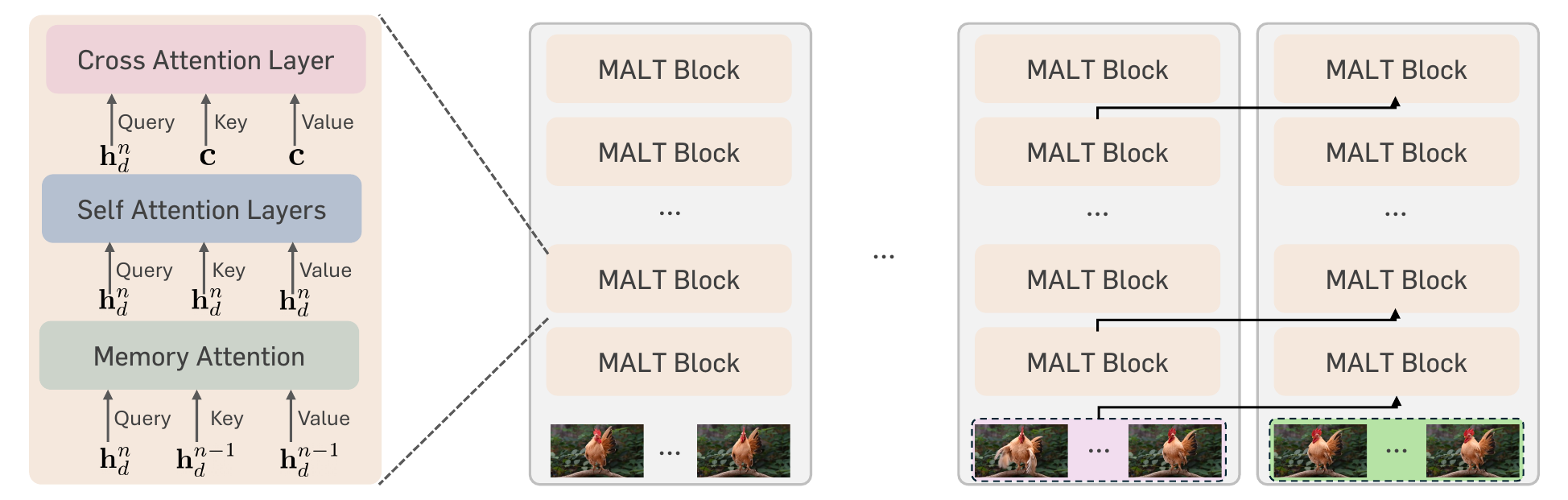}
    \vspace{-0.1in}
    \caption{\textbf{\sname model architecture.} We add memory attention layers to the existing diffusion transformer architectures, which are operated recurrently across segments of short video clips.}
    \label{fig:model}
\end{figure}

To sum up, we train $D_{\bm{\theta}}$ with the following denoising autoencoder objective with the memory $\bh^{n}$:
\begin{align}
    \mathbb{E}
    \Big[
    \lambda(t)||D_{\bm{\theta}}(\bz_t^{n+1}, t;\, \bh^{n}, \bc) - \bz_0^{n+1}||_2^2 
    \Big],
\label{eq:pseudo-obj}
\end{align}
where each data is sampled from the video dataset $\mathcal{D}$, $\epsilon \sim p(\bm{\epsilon})$, $t \sim [1, T]$, and $n \sim P(n)$ with pre-defined prior distributions $p(\bm{\epsilon})$ and $P(n)$. 
Specifically, we set $P(n)$ as $P(n$ {$=$} $0)=\nicefrac{1}{2}$ and $P(n$ {$=$} $k)=\nicefrac{1}{2(N-1)}$ for $k>0$, as generating videos without memory (\ie, $n=0$) is more difficult than continuation with a given memory vector (\ie, $n>0$).
Note that in Eq.~\eqref{eq:pseudo-obj} we do not use the stop-grad operation on $\bh^{n}$ itself; as we mentioned, diffusion model training uses $t \geq 1$ in the common diffusion model objective because they only consider noisy inputs but our memory vector computation uses a clean sample ($t=0$), which cannot be optimized without a backpropagation to $\bh^{n}$. 
Instead, we use the stop-grad operation on the $\bh^{i}$ for $i<n$ which are used to compute $\bh^{n}$ in
order to reduce memory requirements with respect to the number of segments used during training.

\vspace{0.02in}
\noindent\textbf{Training for long term stability.}
Unfortunately, the video quality from the model $D_{\bm{\theta}}$ trained with Eq.~\eqref{eq:pseudo-obj} is usually not satisfactory, because of frame quality degradation caused by error accumulation. We hypothesize that the reason why this happens is because there exists a discrepancy between training and inference: in training, we use ground-truth latent vector $\bz^{n}$ for computation of the memory $\bh^{n}$, but at inference, the model instead uses generated latent vectors, 
where small errors during generation can compound over a long sequence of segments.

To mitigate this discrepancy, we use a noisy version of $\bh^{n}$ at training time, denoted by $\tilde{\bh}^{n}$,  where
\begin{gather}
    \tilde{\bh}^{n} \coloneqq \mathrm{HiddenState}\big(D_{\bm{\theta}} (\bz^{n} + \bm{\xi}, 0;\, \small{\mathtt{sg}}(\bh^{n-1}), \bc)\big),\\ \text{where}\,\,\bh^{0} = [\mathbf{0}, \ldots, \mathbf{0}],
\end{gather}
and $\bm{\xi} \sim p(\bm{\xi})$ with a pre-defined prior distribution $p(\bm{\xi})$. Since $\tilde{\bh}^{n}$ is computed with $\bh^{n-1}$ and a \emph{noisy} latent vector $\bz^{n} + \bm{\xi}$, the model is trained to be robust to small errors and reduces the train-test discrepancy between the memory computed at training and inference. 

Summing up all of these components, our final training objective $\mathcal{L}(\bm{\theta})$ becomes:
\begin{align}
    \mathcal{L}(\bm{\theta}) \coloneqq \mathbb{E} 
    \Big[
    \lambda(t)|| D_{\bm{\theta}}(\bz_t^{n+1}, t;\, \tilde{\bh}^{n}, \bc) - \bz_0^{n+1}||_2^2 
    \Big],
\label{eq:obj}
\end{align}
with prior distributions $p(\bm{\epsilon}), p(n), p(\bm{\xi})$.

For $p(\bm{\epsilon})$, we use a progressively correlated Gaussian distribution proposed in \citet{ge2023preserve} to further mitigate error accumulation. Next, we set $p(\bm{\xi})$ to be a Gaussian $\mathcal{N}(\mathbf{0}$, $\sigma_{\mathrm{mem}}^2\mathbf{I})$ with small $\sigma_{\mathrm{mem}}>0$.

\vspace{0.02in}
\noindent\textbf{Inference.}
After training, we synthesize a long video by autoregressively generating one segment at a time. Specifically, we start from generating a first segment $\bz^{1}$ conditioned on $\bc$, and then iteratively generate $\bz^{n+1}$ for $n>0$ by computing memory $\bh^{n}$ and performing conditional generation from $\bh^{n}$ and $\bc$. We provide detailed pseudocode in Appendix~\ref{appen:sampling}.

\subsection{Architecture}
\label{subsec:arch}

\textbf{Autoencoder.}
Similar to the encoding scheme used in a recent latent video diffusion model, W.A.L.T \citep{gupta2023photorealistic}, we use a
causal 3D CNN encoder-decoder architecture for the video autoencoder based on the  
MAGVIT-2 tokenizer~\citep{yu2024language} without quantization (so that latent vectors lie in a continuous space). We train the autoencoder with a sum of pixel-level reconstruction loss (\eg, mean-squared error), perceptual loss (\eg, LPIPS~\citep{zhang2018perceptual}), and adversarial loss~\citep{goodfellow2014generative} similar to prior image and video generation methods. Recall that both training and inference are not done directly on long videos; they are done after splitting long videos into short segments.

\vspace{0.02in}
\noindent\textbf{Diffusion model.}
As outlined in Figure~\ref{fig:model}, our model architecture is based on the recent diffusion transformer (DiT) architecture~\citep{Peebles2022DiT, ma2024latte, yu2024representation}. 
Thus, given a latent vector $\bz^n \in \mathbb{R}^{l \times h \times w \times c}$ of a video clip $\bx^n$, we patchify it with patch size $p_l \times p_s \times p_s$ to form a flattened latent vector $\mathtt{patchify}(\bz^n) \in \mathbb{R}^{(lhw / p_lp_s^2) \times c}$ with a sequence length $lhw / p_lp_s^2$ and use it as inputs to the model. In particular, we choose W.A.L.T~\citep{gupta2023photorealistic} as backbone, a variant of DiT which employs efficient spatiotemporal windowed attention instead of full attention between large numbers of video patches.

To enable training with long videos with DiT architectures, we introduce a memory-augmented attention layer and insert this layer to the beginning of every Transformer block. Specifically, we design this layer as a cross-attention layer between the previous memory latent vector and the current hidden state, similar to memory-augmented attention~\citep{dai2019transformer} in the language domain. Hence, query, key, and value of a $d$-th memory layer with the segment $\bz^{n}$ and the memory latent vector $\bh^{n-1}$ become:
\begin{gather*}
    \text{query}\coloneqq \bh_{d}^{n},\,\,\text{key} \coloneqq [\bh_{d}^{n-1}, \bh_{d}^{n}],\,\,\text{value}\coloneqq [\bh_{d}^{n-1}, \bh_{d}^{n}], \\ \bh_{d}^{n-1}, \bh_{d}^{n} \in \mathbb{R}^{(hw / p_s^2) \times (l/p_l) \times c'},
\end{gather*}
where $c'$ denotes the hidden dimension of the model and $\bh_{d}^{n-1}, \bh_{d}^{n}$ are \emph{reshaped} as a sequence length $l/p_l$ and a batch dimension size $hw/p_s^2$, similar to previous space-time factorized attention~\citep{bertasius2021space}. 
We also use relative positional encodings that are widely used to handle longer context length.

With this formulation, memory-augmented attentions are only computed together with each of $l/p_l$ patches that have the same spatial location (\ie, temporal attention in video transformers~\citep{bertasius2021space}). This increase is not significant because the computation of attention is restricted only to the sample spatial locations. Our 
memory augmented attentions have $O(L^2HW)$ computational complexity whereas full attention would scale as $O((LHW)^2)$.

Finally, recall that we build our architecture on W.A.L.T, but our general approach of using memory-augmented latent transformers can be applied more broadly to any video diffusion transformer architectures, such as \citet{lu2023vdt} and \citet{ma2024latte}. We provide a detailed illustration of the architecture combined with W.A.L.T in Appendix~\ref{appen:archi}.
\section{Related Work}

We provide a brief discussion with important relevant literature. We provide more detailed discussion in Appendix~\ref{appen:related}. 

\vspace{0.02in}
\noindent\textbf{Video diffusion models.}
Inspired by the remarkable success of image diffusion models~\citep{rombach2021highresolution,chen2023pixart}, many
recent approaches have attempted to solve the challenging problem of video generation through diffusion models by extending ideas and architectures
from the image domain \citep{blattmann2023align,gupta2023photorealistic,ho2022video,harvey2022flexible,he2022lvdm,ge2023preserve,lu2023vdt,ma2024latte,ho2022imagen,singer2022make,voleti2022mcvd,weng2023art,yin2023nuwa,yu2023video,yu2024efficient,zhou2022magicvideo}. Since memory and computation requirements increase 
dramatically due to the cubic complexity of videos as RGB pixels, most works have focused on efficient methods to generate videos via diffusion models. One avenue to achieve efficiency has been to train compact latent representations specialized for videos and training latent diffusion models in such spaces~\citep{gupta2023photorealistic,he2022lvdm,yu2023video,yu2024efficient}. Other works have designed efficient 
model architectures to reduce computation in handling video data~\citep{gupta2023photorealistic,bar2024lumiere,yu2024efficient}. Remarkably, with these efforts, very recent works have shown diffusion models can generate high-resolution (1080p) and quite long (up to 1 minute) videos if they use massive number of videos as data with an extremely large model \citep{veo,videoworldsimulators2024}. Our method builds on both strands of prior work, leading to an
efficient architecture specialized for and trained on \emph{long} videos.

\vspace{0.02in}
\noindent\textbf{Long video generation.}
Many works have attempted to solve long video generation in different directions. First, there exist several works that interpret videos as compactly parameterized continuous functions of time (neural fields; \citep{sitzmann2020implicit}) and train GANs \citep{goodfellow2014generative} to generate such function parameters \citep{yu2022digan,skorokhodov2021stylegan}. These works have shown potential to generate arbitrarily long and smooth videos but are difficult to scale up with complex datasets due to the mode collapse problem of GANs~\citep{srivastava2017veegan}. Another line of work has proposed an autoregressive approach to generate frames conditioned on previous context~\citep{blattmann2023align,gupta2023photorealistic,yan2023temporally,weng2023art,ge2022long,villegas2023phenaki,yan2021videogpt,kondratyuk2023videopoet}. However, due to the high-dimensionality of videos, these approaches condition on a small temporal window of previous frames. They also often suffer from error propagation leading to low-quality frames for longer
temporal horizons~\citep{huang2023video}, particularly on complex datasets.
Several recent works have introduced training (or training-free) method to mitigate this issue \citep{ruhe2024rolling, kim2024fifo, chen2024diffusion, xie2024progressive,lu2024freelong}, but they still do not have long conteuxutal understanding capability.
Other methods have proposed hierarchical generation to progressively generate long videos by interpolating between previously synthesized frames~\citep{yin2023nuwa,huang2023video}. However, if the target video length is extremely long, they need to generate very sparse video frames with very little local correlation, which is challenging, and they are also limited to generating frames up to the training length. \sname can be categorized as a diffusion based
autoregressive approach; however, unlike the above approaches that condition on previously generated frames, we condition on memory, which is capable of capturing information going significantly further back in time, and is robust to error propagation when generating extremely long videos.

\vspace{0.02in}
\noindent\textbf{Transformers for handling long context.}
Recent works primarily from the large language models (LLMs) literature have proposed different ideas to handle extremely long contexts using transformers~\citep{vaswani2017attention}. 
First, several works have introduced efficient attention mechanisms~\citep{wang2020linformer,kitaev2020reformer,beltagy2020longformer,choromanski2020rethinking} that reduce their quadratic complexity and thereby can handle longer contexts better. Next, some approaches have attempted to adopt recurrence in transformers~\citep{dai2019transformer,bessonov2023recurrent,bulatov2022recurrent,bulatov2024beyond,hutchins2022block,peng2023rwkv}, showing great potential to understand long contexts, sometimes even demonstrating their effectiveness on extremely long sequences (1M) \citep{bulatov2024beyond}.
Some works introduce a method to cache previous keys and values from previous contexts (known as \emph{KV-caches})~\citep{wu2022memorizing,adnan2024keyformer,pope2023efficiently}, which are often combined with a relative positional encoding generalizable to a longer sequence than the training length~\citep{su2024roformer}. Finally, there exist some approaches that implement
efficient attention using hardware optimizations \citep{dao2022flashattention,liu2023ring}. Our method fits into this larger family of approaches by incorporating the recurrence technique into recent transformer-based diffusion
approaches~\citep{gupta2023photorealistic,Peebles2022DiT} to generate long sequences. 

\begin{figure*}[t!]
    \centering    
    \includegraphics[width=\textwidth]{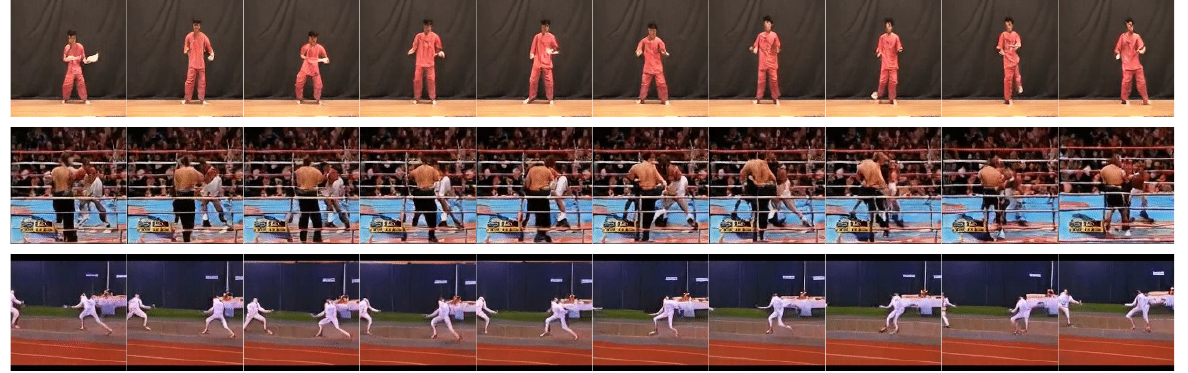}
    \caption{\textbf{Long video prediction results} on UCF-101. Each frame has 128$\times$128 resolution. We visualize the frames with a stride of 16.
    }
    \label{fig:main_qual_ucf101}
\end{figure*}

\section{Experiments}
\begin{figure*}[t!]
    \centering    
    \includegraphics[width=\textwidth]{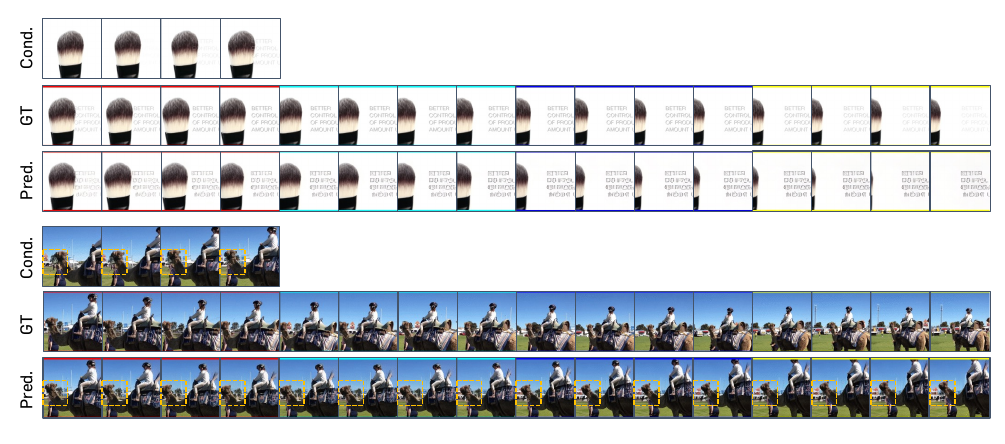}
    \vspace{-0.2in}
    \caption{\textbf{Long video prediction results} on Kinetics-600. Each frame has 128$\times$128 resolution. We visualize the frames with a stride of 5. The first 4 frames indicate an input condition. We mark the predicted frames, where the different color denotes different segments predicted from the model. For each video, we visualize the ground-truth in the first row and the prediction in the second row.
    }
    \label{fig:main_qual_k600}
    \vspace{-0.1in}
\end{figure*}

\begin{figure}[t!]
\centering
    \includegraphics[width=.6\linewidth]{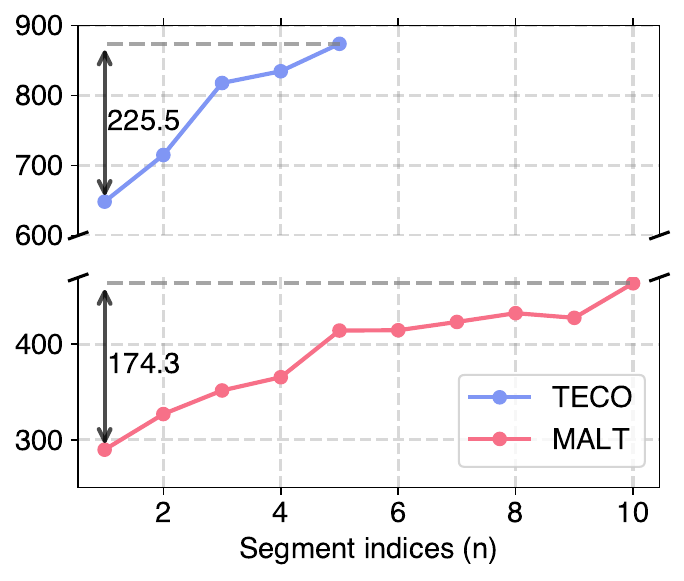}
    \vspace{-0.15in}
    \caption{\textbf{Error propagation analysis.} FVD values on Kinetics-600 measured with a part of segments in the entire videos.}
    \label{fig:analysis}
    \vspace{-0.2in}
\end{figure}

We validate the performance of \sname and the effect of the proposed components through extensive experiments. In particular, we investigate the following questions:
\begin{itemize}[leftmargin=*,itemsep=0mm]
\item Can \sname generate or predict videos with a long horizon with understanding? (Table~\ref{tab:ucf}, \ref{tab:kinetics}, Figure~\ref{fig:main_qual_ucf101}, \ref{fig:main_qual_k600})
\item Do the proposed components (recurrent memory, training for long-term stability) contribute to the final performance of images and videos (Table~\ref{tabs:ablation})?
\item Is the frame quality of generated video maintained across frame indices? (Figure~\ref{fig:analysis}) 
\end{itemize}

\vspace{0.02in}
\noindent\textbf{Note.} Although we achieve strong performance on popular benchmarks, the focus of our paper is more on long-term stability rather than per-frame fidelity. Thus, the size of our models is relatively compact, and we do not train separate super-resolution modules to increase frame quality.

\subsection{Setup}
\label{subsec:setup}

We explain some important setups in this section. We include more details in Appendix~\ref{appen:setup} and \ref{appen:baselines}.

\vspace{0.02in}
\noindent\textbf{Datasets.}
We use long video generation or prediction benchmarks~\citep{skorokhodov2021stylegan, yan2023temporally} to evaluate our method. Specifically, we use UCF-101~\citep{soomro2012ucf101} for unconditional video \emph{generation} with a length of 128 frames and use Kinetics-600~\citep{kay2017kinetics} for long video \emph{prediction} to predict 80 frames conditioned on 20 frames. We choose these video datasets as they contain diverse and complex motions and thus can demonstrate the scalability of \sname to generalize to real-world complex videos. For text-to-video generation, we train on 89M text-short-video
pairs (up to 37 frames) and
970M text-image pairs from public internet and internal
sources. All datasets are center-cropped and resized to 128$\times$128 resolution.

\vspace{0.02in}
\noindent\textbf{Training details.}
All models are trained with the AdamW~\citep{loshchilov2018decoupled} optimizer with a learning rate of 5$e$-5. For our autoencoder, we use the same architecture and configuration as W.A.L.T~\citep{gupta2023photorealistic}. We also use similar diffusion model configurations that W.A.L.T used: specifically, we use the W.A.L.T~XL-config for UCF-101 and W.A.L.T~L-config for Kinetics-600 experiments. We also use the W.A.L.T~L-config for text-to-video experiments. For the memory noise scale $\sigma_{\mathrm{mem}}$, we use 0.1 in all experiments.

\vspace{0.02in}
\noindent\textbf{Metrics.}
We mainly use the Fr\`echet Video Distance (FVD~\citep{unterthiner2018towards}; lower is better) to evaluate the quality of generated videos, following recent works~\citep{yu2023video,skorokhodov2021stylegan,kim2024hybrid}. On Kinetics-600, we also measure Peak signal-to-noise ratio (PSNR; higher is better), SSIM (higher is better), and perceptual metric (LPIPS \citep{zhang2018perceptual}; lower is better) between ground-truth video frames and predicted frames to measure how well the prediction resembles the ground-truth. 

\vspace{0.02in}
\noindent\textbf{Baselines.}
We use existing recent video generation and prediction methods as baselines that are capable of long video generation. Specifically, we use MoCoGAN~\citep{tulyakov2018mocogan}, MoCoGAN-HD~\citep{tian2021good}, DIGAN~\citep{yu2022digan}, StyleGAN-V~\citep{skorokhodov2021stylegan}, PVDM~\citep{yu2023video}, and HVDM~\citep{kim2024hybrid} as baselines for long video generation. For the video prediction task, we consider the following baselines: Perceiver AR~\citep{hawthorne2022general}, Latent FDM~\citep{harvey2022flexible}, and TECO~\citep{yan2023temporally}, which are recent generation methods designed for handling long videos with long-term understanding.

\begin{figure*}[t!]
    \centering    
    \includegraphics[width=\textwidth]{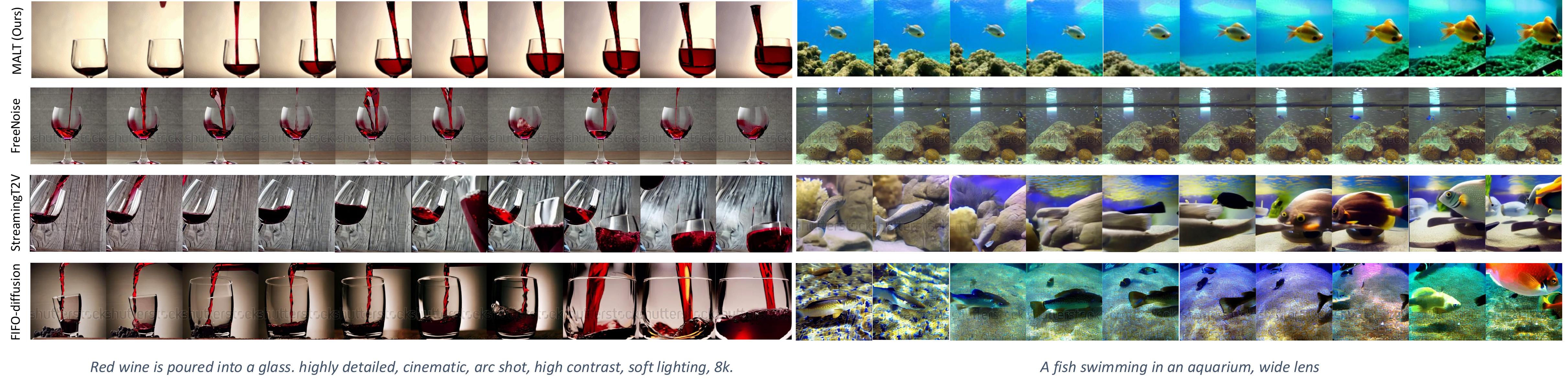}
    \caption{\textbf{Long text-to-video generation results}. We show two generated videos
    denoting frame indices within each frame. Each video 
    is generated at 8 fps spanning $>30$s and each frame has 128$\times$128 resolution.
    Text prompts are provided under each visualization.
    }
    \label{fig:main_qual_t2v}
\end{figure*}

\subsection{Long-term contextual understanding}
First, Figure~\ref{fig:main_qual_ucf101} on UCF-101 shows that our method does understand long-term dependencies through memory latent vectors: in the video of two people fencing (last row), the first person on the left disappears entirely but is able to reappear in the last generated frames.
This is also demonstrated in Figure~\ref{fig:main_qual_k600} where we train on Kinetics-600. For instance, in the second video, there is an orange roof on the left side of the given prefix frames. The roof is covered by the camel's head for a few dozen frames before reappearing in the last predictions. Without a long understanding of the context, the model cannot remember these details, and the final predicted segment cannot contain this roof. 

Note that this is possible due to the fundamental difference between predominant autoregressive methods and \sname: the former is ``explicitly'' autoregressive, while \sname is not. Specifically, prior works generate video frames conditioned on the previous explicit context (\eg,, explicit in the sense of being conditioned on the last 8-16 RGB frames), whereas our method autoregressively generates video frames with respect to a ``latent memory'', which is capable of capturing much longer context in an efficient manner. This allows our model to have an extremely long receptive field without blowing up memory requirements.

\subsection{Long-term stability}
We also quantitatively analyze long term generation stability by visualizing FVD plots of the recent long-video generation training scheme, TECO, and \sname in Figure~\ref{fig:analysis}. Here, FVD values are computed by dividing the predicted video frames into multiple segments of 16-frame video clips and separately measuring FVD values using video clips at each specific interval. As shown in this figure, \sname generates higher-quality videos (\ie, FVD values are always better at any given interval) but also suffers less from error accumulation issues than the previous state-of-the-art TECO. Specifically, the FVD difference between the first and the last FVD value of \sname is smaller (174.3) than TECO (225.5) even if the FVD difference of \sname is measured with 2$\times$ more number of segments (\ie, \sname generates 2$\times$ more video frames). This indicates the frame quality of \sname drops slower than TECO, highlighting the robustness of \sname to error accumulation. 

We also compare the T2V generation results from recent training-free long T2V methods and MALT.\footnote{There are numerous factors that affect the frame quality, to name a few, model size, text caption fidelity, and video-text alignment. Thus, an apples-to-apples comparison of videos based on frame fidelity is difficult because each method uses design choices for aforementioned factors: Thus, we focus on comparing long-term stability of generated video frames.} 
To compare this, we train a relatively small model (545M) compared to existing T2V models that usually have more than 2B parameters \citep{blattmann2023align,gupta2023photorealistic,ho2022imagen}. 
As shown in Figure~\ref{fig:main_qual_t2v}, \sname shows reasonable T2V generation on complex text prompts and more importantly, generated frames are consistent and stable over a long temporal horizon. We provide more visualizations in Appendix~\ref{appen:more_qual}: we show that \sname can be generalized to generate up to 2 minute videos with 8 fps.

\begin{table}[h!]
\centering\small
\caption{
\textbf{System-level comparision.}
(a) FVDs of video generation and models on UCF-101 and (b) FVD, PSNR, SSIM and LPIPS of video prediction models on Kinetics-600. For FVD on Kinetics-600, we measure it with entire video frames, namely including initial ground-truth frames given as contexts and 80 predicted frames from each baseline. For the other three metrics, we measure the values using only the predicted 80 frames. Bold indicate the best results. 
} 
\begin{subtable}{\linewidth}
\subcaption{UCF-101}
\centering\small
\begin{tabular}{l c}
\toprule
Method & {$\text{FVD}$ $\downarrow$} 
\\
\midrule
    MoCoGAN~\citep{tulyakov2018mocogan}         
    & 3679.0 \\
    + StyleGAN2~\citep{karras2020analyzing}     
    & 2311.3 \\
    MoCoGAN-HD~\citep{tian2021good}      
    & 2606.5 \\
    DIGAN~\citep{yu2022digan}           
    & 2293.7 \\
    StyleGAN-V~\citep{skorokhodov2021stylegan}      
    & 1773.4 \\
    Latte \citep{ma2024latte} + FreeNoise \citep{qiu2023freenoise}  &  1157.7 \\
    MeBT \citep{yoo2023towards} & \phantom{0}{968}\phantom{.0} \\
    PVDM~\citep{yu2023video} & \phantom{0}{505.0} \\
    HVDM~\citep{kim2024hybrid} & \phantom{0}{549.7} \\
    CoordTok \citep{jang2024efficient} + SiT-L/2 \citep{ma2024sit}  & \phantom{0}369.3 \\
    \midrule
    \textbf{\sname (Ours)} & \phantom{0}{\textbf{220.4}} \\
    \bottomrule
\end{tabular}
\label{tab:ucf}
\end{subtable}
\begin{subtable}{\linewidth}
\vspace{0.07in}
\subcaption{Kinetics-600}
\centering\small
\begin{tabular}{l cccc}
    \toprule
    Method & FVD $\downarrow$ & PSNR $\uparrow$ & SSIM $\uparrow$ & LPIPS $\downarrow$ \\
    \midrule
    Perceiver AR~\citep{hawthorne2022general}
    & 1022 & 13.4 & 0.310 & 0.404
    \\
    Latent FDM~\citep{harvey2022flexible}
    & \phantom{0}{960} & 13.2 & 0.334 & 0.413 \\ 
    TECO~\citep{yan2023temporally}
    & \phantom{0}{799} & 13.8 & 0.341 & 0.381 \\
    \midrule
    \textbf{\sname (Ours)} & 
    \phantom{0}{\textbf{392}} & \textbf{15.4} & \textbf{0.437} & \textbf{0.276} \\
    \bottomrule
\end{tabular}
\label{tab:kinetics}
\end{subtable}
\end{table}

\subsection{System-level comparison}
Table~\ref{tab:ucf} and \ref{tab:kinetics} summarize long video generation and prediction results on UCF-101 and Kinetics-600 datasets, respectively. On UCF-101, \sname achieves an FVD score (lower is better) of 220.4, significantly outperforming existing video generation methods including unconditional latent video diffusion models capable of long video generation, such as PVDM~\citep{yu2023video} and HVDM~\citep{kim2024hybrid}. Similarly, on a long video prediction task on Kinetics-600, \sname shows much better performance than the previous state-of-the-art method, TECO, on all four metrics despite have $\sim 2\times $ fewer parameters (440M for \sname and 1.1B for TECO).

\newcommand{\recurrentg}{\textcolor{lightgray}{Recurrent}\xspace}
\newcommand{\memvitxlg}{\textcolor{lightgray}{\sname-L}\xspace}%
\newcommand{\fiveoneoneg}{\textcolor{lightgray}{(5, 1, 1)}\xspace}%
\newcommand{\zeroponeg}{\textcolor{lightgray}{\checkmark}\xspace}%
\newcommand{\twentyg}{\textcolor{lightgray}{4}\xspace}%
\newcommand{\upweightedg}{\textcolor{lightgray}{Upweighted}\xspace}%
\begin{table}[ht!]
    \caption{\textbf{Ablation studies.} FVD, PSNR, and SSIM values on Kinetics-600. Gray indicates that the component is fixed for ablation.``Last only'' a condition model only on the last segment. ``kv cache'': a conditional model that predicts a segment on all previous segments without any compression. ``Recurrent'': our recurrent memory scheme that uses fixed-size hidden states as condition.}
    \vspace{-0.05in}
    \centering\small
    \begin{tabular}{l c c c c c c c c}
    \toprule
    & {} & {Robust}  & FVD  & PSNR & SSIM\\
    \midrule
    \rowcolor{gray!30}
    \multicolumn{2}{c}{W.A.L.T-L \citep{gupta2023photorealistic}}  
     & {} & 459 & 14.8 & 0.400 \\
    \midrule
    \multirow{3}{*}{\shortstack[l]{Memory \\ design}}  
    & {Last only} & {\zeroponeg} & 423 & 15.0 & 0.424 \\
    & {kv-cache}  & {\zeroponeg} & 396 & 15.1 & 0.429   \\
    & {Recurrent} & {\zeroponeg} & 392 & 15.4 & 0.437  \\
    \midrule
    \multirow{2}{*}{\shortstack[l]{Training \\ technique}}
    & {\recurrentg}  & {} & {383} & 14.7 & 0.401 \\
    & {\recurrentg}  & {\checkmark} & 392 & 15.4 & 0.437  \\
    \bottomrule
    \end{tabular}
    \label{tabs:ablation}
\end{table}


\subsection{Ablation studies}
\label{subsec:ablation}

We conduct ablation studies and report in Table~\ref{tabs:ablation}. We also perform error propagation analysis and summarize the result in Figure~\ref{fig:analysis}. For further analysis, see Appendix~\ref{appen:ablation}.

\vspace{0.02in}
\noindent\textbf{Memory design.}
As shown in Table~\ref{tabs:ablation}, a conditional model that only uses the last segment (``Last only'') shows worse performance than other models that use longer contexts, indicating the importance of long-term contextual understanding. Moreover, our recurrent and compact memory design (``Recurrent'') shows comparable performance compared to the kv-cache that uses the entire context without any compression, which speaks to the effectiveness of \sname.

\vspace{0.02in}
\noindent\textbf{Training objective.}
The model trained without noise augmentation and correlated prior distribution shows worse SSIM and PSNR (\eg, PSNR gets worse from 15.4 to 14.7), exhibiting the frame quality drop across the frame indices. It demonstrates the importance of our training techniques that use noise augmentation to the memory latent vector and correlated prior distribution for stabilizing the frame quality.

\section{Conclusion}
\label{sec:conclusion}
We proposed \sname Diffusion, a latent video diffusion model for any-length video generation. It is based on proposing a new DiT architecture, which employs a memory for encoding long video context into succinct low-dimensional latent vectors. Our approach is general and can be applied in principle to any architectures and we believe it will facilitate long video synthesis breaking beyond the current limits.
We discuss social impact and limitations in Appendix \ref{appen:social_impact}.

\section*{Acknowledgments}
We thank Matt Falkenhagen for his assistance with this project. SY also thanks Younggyo Seo and Changyeon Kim for their insightful comments, as well as Gwanghyun Kim for helpful discussions on diffusion model implementation.

{
    \small
    \bibliographystyle{ieeenat_fullname}
    \bibliography{references}

\begin{thebibliography}{122}
\providecommand{\natexlab}[1]{#1}
\providecommand{\url}[1]{\texttt{#1}}
\expandafter\ifx\csname urlstyle\endcsname\relax
  \providecommand{\doi}[1]{doi: #1}\else
  \providecommand{\doi}{doi: \begingroup \urlstyle{rm}\Url}\fi

\bibitem[Acharya et~al.(2018)Acharya, Huang, Paudel, and Van~Gool]{acharya2018towards}
Dinesh Acharya, Zhiwu Huang, Danda~Pani Paudel, and Luc Van~Gool.
\newblock Towards high resolution video generation with progressive growing of sliced {Wasserstein GAN}s.
\newblock \emph{arXiv preprint arXiv:1810.02419}, 2018.

\bibitem[Adnan et~al.(2024)Adnan, Arunkumar, Jain, Nair, Soloveychik, and Kamath]{adnan2024keyformer}
Muhammad Adnan, Akhil Arunkumar, Gaurav Jain, Prashant~J Nair, Ilya Soloveychik, and Purushotham Kamath.
\newblock Keyformer: Kv cache reduction through key tokens selection for efficient generative inference.
\newblock \emph{arXiv preprint arXiv:2403.09054}, 2024.

\bibitem[An et~al.(2023)An, Zhang, Yang, Gupta, Huang, Luo, and Yin]{an2023latent}
Jie An, Songyang Zhang, Harry Yang, Sonal Gupta, Jia-Bin Huang, Jiebo Luo, and Xi Yin.
\newblock Latent-shift: Latent diffusion with temporal shift for efficient text-to-video generation.
\newblock \emph{arXiv preprint arXiv:2304.08477}, 2023.

\bibitem[Babaeizadeh et~al.(2018)Babaeizadeh, Finn, Erhan, Campbell, and Levine]{babaeizadeh2018stochastic}
Mohammad Babaeizadeh, Chelsea Finn, Dumitru Erhan, Roy~H Campbell, and Sergey Levine.
\newblock Stochastic variational video prediction.
\newblock In \emph{International Conference on Learning Representations}, 2018.

\bibitem[Bar-Tal et~al.(2024)Bar-Tal, Chefer, Tov, Herrmann, Paiss, Zada, Ephrat, Hur, Li, Michaeli, et~al.]{bar2024lumiere}
Omer Bar-Tal, Hila Chefer, Omer Tov, Charles Herrmann, Roni Paiss, Shiran Zada, Ariel Ephrat, Junhwa Hur, Yuanzhen Li, Tomer Michaeli, et~al.
\newblock Lumiere: A space-time diffusion model for video generation.
\newblock \emph{arXiv preprint arXiv:2401.12945}, 2024.

\bibitem[Beltagy et~al.(2020)Beltagy, Peters, and Cohan]{beltagy2020longformer}
Iz Beltagy, Matthew~E Peters, and Arman Cohan.
\newblock Longformer: The long-document transformer.
\newblock \emph{arXiv preprint arXiv:2004.05150}, 2020.

\bibitem[Bertasius et~al.(2021)Bertasius, Wang, and Torresani]{bertasius2021space}
Gedas Bertasius, Heng Wang, and Lorenzo Torresani.
\newblock Is space-time attention all you need for video understanding?
\newblock In \emph{International Conference on Machine Learning}, 2021.

\bibitem[Bessonov et~al.(2023)Bessonov, Staroverov, Zhang, Kovalev, Yudin, and Panov]{bessonov2023recurrent}
Arkadii Bessonov, Alexey Staroverov, Huzhenyu Zhang, Alexey~K Kovalev, Dmitry Yudin, and Aleksandr~I Panov.
\newblock Recurrent memory decision transformer.
\newblock \emph{arXiv preprint arXiv:2306.09459}, 2023.

\bibitem[Blattmann et~al.(2023)Blattmann, Rombach, Ling, Dockhorn, Kim, Fidler, and Kreis]{blattmann2023align}
Andreas Blattmann, Robin Rombach, Huan Ling, Tim Dockhorn, Seung~Wook Kim, Sanja Fidler, and Karsten Kreis.
\newblock Align your latents: High-resolution video synthesis with latent diffusion models.
\newblock In \emph{IEEE Conference on Computer Vision and Pattern Recognition}, 2023.

\bibitem[Brooks et~al.(2024)Brooks, Peebles, Holmes, DePue, Guo, Jing, Schnurr, Taylor, Luhman, Luhman, Ng, Wang, and Ramesh]{videoworldsimulators2024}
Tim Brooks, Bill Peebles, Connor Holmes, Will DePue, Yufei Guo, Li Jing, David Schnurr, Joe Taylor, Troy Luhman, Eric Luhman, Clarence Ng, Ricky Wang, and Aditya Ramesh.
\newblock Video generation models as world simulators.
\newblock \emph{OpenAI Blog}, 2024.

\bibitem[Bulatov et~al.(2022)Bulatov, Kuratov, and Burtsev]{bulatov2022recurrent}
Aydar Bulatov, Yury Kuratov, and Mikhail Burtsev.
\newblock Recurrent memory transformer.
\newblock In \emph{Advances in Neural Information Processing Systems}, 2022.

\bibitem[Bulatov et~al.(2024)Bulatov, Kuratov, Kapushev, and Burtsev]{bulatov2024beyond}
Aydar Bulatov, Yuri Kuratov, Yermek Kapushev, and Mikhail Burtsev.
\newblock Beyond attention: Breaking the limits of transformer context length with recurrent memory.
\newblock In \emph{AAAI Conference on Artificial Intelligence}, 2024.

\bibitem[Chang et~al.(2022)Chang, Zhang, Jiang, Liu, and Freeman]{chang2022maskgit}
Huiwen Chang, Han Zhang, Lu Jiang, Ce Liu, and William~T Freeman.
\newblock Maskgit: Masked generative image transformer.
\newblock In \emph{IEEE Conference on Computer Vision and Pattern Recognition}, 2022.

\bibitem[Chen et~al.(2024{\natexlab{a}})Chen, Monso, Du, Simchowitz, Tedrake, and Sitzmann]{chen2024diffusion}
Boyuan Chen, Diego~Marti Monso, Yilun Du, Max Simchowitz, Russ Tedrake, and Vincent Sitzmann.
\newblock Diffusion forcing: Next-token prediction meets full-sequence diffusion.
\newblock \emph{arXiv preprint arXiv:2407.01392}, 2024{\natexlab{a}}.

\bibitem[Chen et~al.(2024{\natexlab{b}})Chen, Yu, Ge, Yao, Xie, Wu, Wang, Kwok, Luo, Lu, et~al.]{chen2023pixart}
Junsong Chen, Jincheng Yu, Chongjian Ge, Lewei Yao, Enze Xie, Yue Wu, Zhongdao Wang, James Kwok, Ping Luo, Huchuan Lu, et~al.
\newblock Pixart-$\alpha$: Fast training of diffusion transformer for photorealistic text-to-image synthesis.
\newblock In \emph{International Conference on Learning Representations}, 2024{\natexlab{b}}.

\bibitem[Choromanski et~al.(2020)Choromanski, Likhosherstov, Dohan, Song, Gane, Sarlos, Hawkins, Davis, Mohiuddin, Kaiser, et~al.]{choromanski2020rethinking}
Krzysztof Choromanski, Valerii Likhosherstov, David Dohan, Xingyou Song, Andreea Gane, Tamas Sarlos, Peter Hawkins, Jared Davis, Afroz Mohiuddin, Lukasz Kaiser, et~al.
\newblock Rethinking attention with performers.
\newblock \emph{arXiv preprint arXiv:2009.14794}, 2020.

\bibitem[Clark et~al.(2019)Clark, Donahue, and Simonyan]{clark2019adversarial}
Aidan Clark, Jeff Donahue, and Karen Simonyan.
\newblock Adversarial video generation on complex datasets.
\newblock \emph{arXiv preprint arXiv:1907.06571}, 2019.

\bibitem[Dai et~al.(2019)Dai, Yang, Yang, Carbonell, Le, and Salakhutdinov]{dai2019transformer}
Zihang Dai, Zhilin Yang, Yiming Yang, Jaime~G Carbonell, Quoc Le, and Ruslan Salakhutdinov.
\newblock Transformer-xl: Attentive language models beyond a fixed-length context.
\newblock In \emph{Annual Meeting of the Association for Computational Linguistics}, 2019.

\bibitem[Dao et~al.(2022)Dao, Fu, Ermon, Rudra, and R{\'e}]{dao2022flashattention}
Tri Dao, Dan Fu, Stefano Ermon, Atri Rudra, and Christopher R{\'e}.
\newblock Flashattention: Fast and memory-efficient exact attention with io-awareness.
\newblock In \emph{Advances in Neural Information Processing Systems}, 2022.

\bibitem[DeepMind(2024)]{veo}
Google DeepMind.
\newblock Veo: Our most capable generative video model.
\newblock \emph{Google DeepMind Technologies}, 2024.

\bibitem[Denton and Birodkar(2017)]{denton2017unsupervised}
Emily Denton and Vighnesh Birodkar.
\newblock Unsupervised learning of disentangled representations from video.
\newblock In \emph{Advances in Neural Information Processing Systems}, 2017.

\bibitem[Denton and Fergus(2018)]{denton2018stochastic}
Emily Denton and Rob Fergus.
\newblock Stochastic video generation with a learned prior.
\newblock In \emph{International Conference on Machine Learning}, 2018.

\bibitem[Dhariwal and Nichol(2021)]{dhariwal2021diffusion}
Prafulla Dhariwal and Alex Nichol.
\newblock Diffusion models beat {GAN}s on image synthesis.
\newblock In \emph{Advances in Neural Information Processing Systems}, 2021.

\bibitem[Finn et~al.(2016)Finn, Goodfellow, and Levine]{finn2016unsupervised}
Chelsea Finn, Ian Goodfellow, and Sergey Levine.
\newblock Unsupervised learning for physical interaction through video prediction.
\newblock In \emph{Advances in Neural Information Processing Systems}, 2016.

\bibitem[Fox et~al.(2021)Fox, Tewari, Elgharib, and Theobalt]{fox2021stylevideogan}
Gereon Fox, Ayush Tewari, Mohamed Elgharib, and Christian Theobalt.
\newblock {StyleVideoGAN}: A temporal generative model using a pretrained {StyleGAN}.
\newblock \emph{arXiv preprint arXiv:2107.07224}, 2021.

\bibitem[Franceschi et~al.(2020)Franceschi, Delasalles, Chen, Lamprier, and Gallinari]{franceschi2020stochastic}
Jean-Yves Franceschi, Edouard Delasalles, Micka{\"e}l Chen, Sylvain Lamprier, and Patrick Gallinari.
\newblock Stochastic latent residual video prediction.
\newblock In \emph{International Conference on Machine Learning}, 2020.

\bibitem[Ge et~al.(2022)Ge, Hayes, Yang, Yin, Pang, Jacobs, Huang, and Parikh]{ge2022long}
Songwei Ge, Thomas Hayes, Harry Yang, Xi Yin, Guan Pang, David Jacobs, Jia-Bin Huang, and Devi Parikh.
\newblock Long video generation with time-agnostic {VQGAN} and time-sensitive transformer.
\newblock In \emph{European Conference on Computer Vision}, 2022.

\bibitem[Ge et~al.(2023)Ge, Nah, Liu, Poon, Tao, Catanzaro, Jacobs, Huang, Liu, and Balaji]{ge2023preserve}
Songwei Ge, Seungjun Nah, Guilin Liu, Tyler Poon, Andrew Tao, Bryan Catanzaro, David Jacobs, Jia-Bin Huang, Ming-Yu Liu, and Yogesh Balaji.
\newblock Preserve your own correlation: A noise prior for video diffusion models.
\newblock In \emph{IEEE International Conference on Computer Vision}, 2023.

\bibitem[Goodfellow et~al.(2014)Goodfellow, Pouget-Abadie, Mirza, Xu, Warde-Farley, Ozair, Courville, and Bengio]{goodfellow2014generative}
Ian Goodfellow, Jean Pouget-Abadie, Mehdi Mirza, Bing Xu, David Warde-Farley, Sherjil Ozair, Aaron Courville, and Yoshua Bengio.
\newblock Generative adversarial nets.
\newblock In \emph{Advances in Neural Information Processing Systems}, 2014.

\bibitem[Gordon and Parde(2021)]{gordon2021latent}
Cade Gordon and Natalie Parde.
\newblock Latent neural differential equations for video generation.
\newblock In \emph{NeurIPS 2020 Workshop on Pre-registration in Machine Learning}, 2021.

\bibitem[Gupta et~al.(2023)Gupta, Yu, Sohn, Gu, Hahn, Fei-Fei, Essa, Jiang, and Lezama]{gupta2023photorealistic}
Agrim Gupta, Lijun Yu, Kihyuk Sohn, Xiuye Gu, Meera Hahn, Li Fei-Fei, Irfan Essa, Lu Jiang, and Jos{\'e} Lezama.
\newblock Photorealistic video generation with diffusion models.
\newblock \emph{arXiv preprint arXiv:2312.06662}, 2023.

\bibitem[Harvey et~al.(2022)Harvey, Naderiparizi, Masrani, Weilbach, and Wood]{harvey2022flexible}
William Harvey, Saeid Naderiparizi, Vaden Masrani, Christian Weilbach, and Frank Wood.
\newblock Flexible diffusion modeling of long videos.
\newblock In \emph{Advances in Neural Information Processing Systems}, 2022.

\bibitem[Hawthorne et~al.(2022)Hawthorne, Jaegle, Cangea, Borgeaud, Nash, Malinowski, Dieleman, Vinyals, Botvinick, Simon, et~al.]{hawthorne2022general}
Curtis Hawthorne, Andrew Jaegle, C{\u{a}}t{\u{a}}lina Cangea, Sebastian Borgeaud, Charlie Nash, Mateusz Malinowski, Sander Dieleman, Oriol Vinyals, Matthew Botvinick, Ian Simon, et~al.
\newblock General-purpose, long-context autoregressive modeling with perceiver ar.
\newblock In \emph{International Conference on Machine Learning}, 2022.

\bibitem[He et~al.(2022)He, Yang, Zhang, Shan, and Chen]{he2022lvdm}
Yingqing He, Tianyu Yang, Yong Zhang, Ying Shan, and Qifeng Chen.
\newblock Latent video diffusion models for high-fidelity video generation with arbitrary lengths.
\newblock \emph{arXiv preprint arXiv:2211.13221}, 2022.

\bibitem[Ho et~al.(2020)Ho, Jain, and Abbeel]{ho2021denoising}
Jonathan Ho, Ajay Jain, and Pieter Abbeel.
\newblock Denoising diffusion probabilistic models.
\newblock In \emph{Advances in Neural Information Processing Systems}, 2020.

\bibitem[Ho et~al.(2022{\natexlab{a}})Ho, Chan, Saharia, Whang, Gao, Gritsenko, Kingma, Poole, Norouzi, Fleet, and Salimans]{ho2022imagen}
Jonathan Ho, William Chan, Chitwan Saharia, Jay Whang, Ruiqi Gao, Alexey Gritsenko, Diederik~P. Kingma, Ben Poole, Mohammad Norouzi, David~J. Fleet, and Tim Salimans.
\newblock Imagen video: High definition video generation with diffusion models.
\newblock \emph{arXiv preprint arXiv:2210.02303}, 2022{\natexlab{a}}.

\bibitem[Ho et~al.(2022{\natexlab{b}})Ho, Salimans, Gritsenko, Chan, Norouzi, and Fleet]{ho2022video}
Jonathan Ho, Tim Salimans, Alexey Gritsenko, William Chan, Mohammad Norouzi, and David~J Fleet.
\newblock Video diffusion models.
\newblock In \emph{Advances in Neural Information Processing Systems}, 2022{\natexlab{b}}.

\bibitem[H{\"o}ppe et~al.(2022)H{\"o}ppe, Mehrjou, Bauer, Nielsen, and Dittadi]{hoppe2022diffusion}
Tobias H{\"o}ppe, Arash Mehrjou, Stefan Bauer, Didrik Nielsen, and Andrea Dittadi.
\newblock Diffusion models for video prediction and infilling.
\newblock \emph{Transactions on Machine Learning Research}, 2022.

\bibitem[Hu et~al.(2021)Hu, Shen, Wallis, Allen-Zhu, Li, Wang, Wang, and Chen]{hu2021lora}
Edward~J Hu, Yelong Shen, Phillip Wallis, Zeyuan Allen-Zhu, Yuanzhi Li, Shean Wang, Lu Wang, and Weizhu Chen.
\newblock Lora: Low-rank adaptation of large language models.
\newblock \emph{arXiv preprint arXiv:2106.09685}, 2021.

\bibitem[Huang et~al.(2023)Huang, Su, and Yang]{huang2023video}
Hsin-Ping Huang, Yu-Chuan Su, and Ming-Hsuan Yang.
\newblock Video generation beyond a single clip.
\newblock \emph{arXiv preprint arXiv:2304.07483}, 2023.

\bibitem[Hutchins et~al.(2022)Hutchins, Schlag, Wu, Dyer, and Neyshabur]{hutchins2022block}
DeLesley Hutchins, Imanol Schlag, Yuhuai Wu, Ethan Dyer, and Behnam Neyshabur.
\newblock Block-recurrent transformers.
\newblock In \emph{Advances in Neural Information Processing Systems}, 2022.

\bibitem[Jang et~al.(2024)Jang, Yu, Shin, Abbeel, and Seo]{jang2024efficient}
Huiwon Jang, Sihyun Yu, Jinwoo Shin, Pieter Abbeel, and Younggyo Seo.
\newblock Efficient long video tokenization via coordinated-based patch reconstruction.
\newblock \emph{arXiv preprint arXiv:2411.14762}, 2024.

\bibitem[Kahembwe and Ramamoorthy(2020)]{kahembwe2020lower}
Emmanuel Kahembwe and Subramanian Ramamoorthy.
\newblock Lower dimensional kernels for video discriminators.
\newblock \emph{Neural Networks}, 132:\penalty0 506--520, 2020.

\bibitem[Kalchbrenner et~al.(2017)Kalchbrenner, Oord, Simonyan, Danihelka, Vinyals, Graves, and Kavukcuoglu]{kalchbrenner2017video}
Nal Kalchbrenner, A{\"a}ron Oord, Karen Simonyan, Ivo Danihelka, Oriol Vinyals, Alex Graves, and Koray Kavukcuoglu.
\newblock Video pixel networks.
\newblock In \emph{International Conference on Machine Learning}, 2017.

\bibitem[Karras et~al.(2020)Karras, Laine, Aittala, Hellsten, Lehtinen, and Aila]{karras2020analyzing}
Tero Karras, Samuli Laine, Miika Aittala, Janne Hellsten, Jaakko Lehtinen, and Timo Aila.
\newblock Analyzing and improving the image quality of {StyleGAN}.
\newblock In \emph{IEEE Conference on Computer Vision and Pattern Recognition}, 2020.

\bibitem[Karras et~al.(2022)Karras, Aittala, Aila, and Laine]{karras2022edm}
Tero Karras, Miika Aittala, Timo Aila, and Samuli Laine.
\newblock Elucidating the design space of diffusion-based generative models.
\newblock In \emph{Advances in Neural Information Processing Systems}, 2022.

\bibitem[Kay et~al.(2017)Kay, Carreira, Simonyan, Zhang, Hillier, Vijayanarasimhan, Viola, Green, Back, Natsev, et~al.]{kay2017kinetics}
Will Kay, Joao Carreira, Karen Simonyan, Brian Zhang, Chloe Hillier, Sudheendra Vijayanarasimhan, Fabio Viola, Tim Green, Trevor Back, Paul Natsev, et~al.
\newblock The kinetics human action video dataset.
\newblock \emph{arXiv preprint arXiv:1705.06950}, 2017.

\bibitem[Kim et~al.(2024{\natexlab{a}})Kim, Kang, Choi, and Han]{kim2024fifo}
Jihwan Kim, Junoh Kang, Jinyoung Choi, and Bohyung Han.
\newblock Fifo-diffusion: Generating infinite videos from text without training.
\newblock \emph{arXiv preprint arXiv:2405.11473}, 2024{\natexlab{a}}.

\bibitem[Kim et~al.(2024{\natexlab{b}})Kim, Lee, Park, Kim, Lee, Kim, and Yoo]{kim2024hybrid}
Kihong Kim, Haneol Lee, Jihye Park, Seyeon Kim, Kwanghee Lee, Seungryong Kim, and Jaejun Yoo.
\newblock Hybrid video diffusion models with 2d triplane and 3d wavelet representation.
\newblock \emph{arXiv preprint arXiv:2402.13729}, 2024{\natexlab{b}}.

\bibitem[Kim et~al.(2022)Kim, Yu, Lee, and Shin]{kim2022scalable}
Subin Kim, Sihyun Yu, Jaeho Lee, and Jinwoo Shin.
\newblock Scalable neural video representations with learnable positional features.
\newblock In \emph{Advances in Neural Information Processing Systems}, 2022.

\bibitem[Kitaev et~al.(2020)Kitaev, Kaiser, and Levskaya]{kitaev2020reformer}
Nikita Kitaev, {\L}ukasz Kaiser, and Anselm Levskaya.
\newblock Reformer: The efficient transformer.
\newblock In \emph{International Conference on Learning Representations}, 2020.

\bibitem[Kondratyuk et~al.(2023)Kondratyuk, Yu, Gu, Lezama, Huang, Hornung, Adam, Akbari, Alon, Birodkar, et~al.]{kondratyuk2023videopoet}
Dan Kondratyuk, Lijun Yu, Xiuye Gu, Jos{\'e} Lezama, Jonathan Huang, Rachel Hornung, Hartwig Adam, Hassan Akbari, Yair Alon, Vighnesh Birodkar, et~al.
\newblock Videopoet: A large language model for zero-shot video generation.
\newblock \emph{arXiv preprint arXiv:2312.14125}, 2023.

\bibitem[Kong et~al.(2021)Kong, Ping, Huang, Zhao, and Catanzaro]{kong2020diffwave}
Zhifeng Kong, Wei Ping, Jiaji Huang, Kexin Zhao, and Bryan Catanzaro.
\newblock Diffwave: A versatile diffusion model for audio synthesis.
\newblock In \emph{International Conference on Learning Representations}, 2021.

\bibitem[Kumar et~al.(2020)Kumar, Babaeizadeh, Erhan, Finn, Levine, Dinh, and Kingma]{kumar2020videoflow}
Manoj Kumar, Mohammad Babaeizadeh, Dumitru Erhan, Chelsea Finn, Sergey Levine, Laurent Dinh, and Durk Kingma.
\newblock Videoflow: A conditional flow-based model for stochastic video generation.
\newblock In \emph{International Conference on Learning Representations}, 2020.

\bibitem[Lakhotia et~al.(2021)Lakhotia, Kharitonov, Hsu, Adi, Polyak, Bolte, Nguyen, Copet, Baevski, Mohamed, et~al.]{lakhotia2021generative}
Kushal Lakhotia, Evgeny Kharitonov, Wei-Ning Hsu, Yossi Adi, Adam Polyak, Benjamin Bolte, Tu-Anh Nguyen, Jade Copet, Alexei Baevski, Adelrahman Mohamed, et~al.
\newblock Generative spoken language modeling from raw audio.
\newblock \emph{arXiv preprint arXiv:2102.01192}, 2021.

\bibitem[Lee et~al.(2018)Lee, Zhang, Ebert, Abbeel, Finn, and Levine]{lee2018stochastic}
Alex~X Lee, Richard Zhang, Frederik Ebert, Pieter Abbeel, Chelsea Finn, and Sergey Levine.
\newblock Stochastic adversarial video prediction.
\newblock \emph{arXiv preprint arXiv:1804.01523}, 2018.

\bibitem[Lee et~al.(2024)Lee, Sohn, and Shin]{lee2024dreamflow}
Kyungmin Lee, Kihyuk Sohn, and Jinwoo Shin.
\newblock Dreamflow: High-quality text-to-3d generation by approximating probability flow.
\newblock In \emph{International Conference on Learning Representations}, 2024.

\bibitem[Lee et~al.(2021)Lee, Jung, Zhang, Chen, Koh, Huang, Yoon, Lee, and Hong]{lee2021revisiting}
Wonkwang Lee, Whie Jung, Han Zhang, Ting Chen, Jing~Yu Koh, Thomas Huang, Hyungsuk Yoon, Honglak Lee, and Seunghoon Hong.
\newblock Revisiting hierarchical approach for persistent long-term video prediction.
\newblock In \emph{International Conference on Learning Representations}, 2021.

\bibitem[Lippe et~al.(2023)Lippe, Veeling, Perdikaris, Turner, and Brandstetter]{lippe2023pderefiner}
Phillip Lippe, Bastiaan~S. Veeling, Paris Perdikaris, Richard~E Turner, and Johannes Brandstetter.
\newblock {PDE}-refiner: Achieving accurate long rollouts with neural {PDE} solvers.
\newblock In \emph{International Conference on Learning Representations}, 2023.

\bibitem[Liu et~al.(2023)Liu, Zaharia, and Abbeel]{liu2023ring}
Hao Liu, Matei Zaharia, and Pieter Abbeel.
\newblock Ring attention with blockwise transformers for near-infinite context.
\newblock In \emph{Advances in Neural Information Processing Systems}, 2023.

\bibitem[Loshchilov and Hutter(2019)]{loshchilov2018decoupled}
Ilya Loshchilov and Frank Hutter.
\newblock Decoupled weight decay regularization.
\newblock In \emph{International Conference on Learning Representations}, 2019.

\bibitem[Lu et~al.(2023)Lu, Yang, Fei, Huo, Lu, Luo, and Ding]{lu2023vdt}
Haoyu Lu, Guoxing Yang, Nanyi Fei, Yuqi Huo, Zhiwu Lu, Ping Luo, and Mingyu Ding.
\newblock Vdt: An empirical study on video diffusion with transformers.
\newblock \emph{arXiv preprint arXiv:2305.13311}, 2023.

\bibitem[Lu et~al.(2024{\natexlab{a}})Lu, CAI, Li, Qin, and Li]{lu2024improve}
Kexin Lu, Yuxi CAI, Lan Li, Dafei Qin, and Guodong Li.
\newblock Improve temporal consistency in diffusion models through noise correlations, 2024{\natexlab{a}}.

\bibitem[Lu et~al.(2024{\natexlab{b}})Lu, Liang, Zhu, and Yang]{lu2024freelong}
Yu Lu, Yuanzhi Liang, Linchao Zhu, and Yi Yang.
\newblock Freelong: Training-free long video generation with spectralblend temporal attention.
\newblock \emph{arXiv preprint arXiv:2407.19918}, 2024{\natexlab{b}}.

\bibitem[Luc et~al.(2020)Luc, Clark, Dieleman, Casas, Doron, Cassirer, and Simonyan]{luc2020transformation}
Pauline Luc, Aidan Clark, Sander Dieleman, Diego de~Las Casas, Yotam Doron, Albin Cassirer, and Karen Simonyan.
\newblock Transformation-based adversarial video prediction on large-scale data.
\newblock \emph{arXiv preprint arXiv:2003.04035}, 2020.

\bibitem[Luo and Hu(2021)]{luo2021diffusion}
Shitong Luo and Wei Hu.
\newblock Diffusion probabilistic models for 3d point cloud generation.
\newblock In \emph{IEEE Conference on Computer Vision and Pattern Recognition}, 2021.

\bibitem[Ma et~al.(2024{\natexlab{a}})Ma, Goldstein, Albergo, Boffi, Vanden-Eijnden, and Xie]{ma2024sit}
Nanye Ma, Mark Goldstein, Michael~S Albergo, Nicholas~M Boffi, Eric Vanden-Eijnden, and Saining Xie.
\newblock {SiT}: Exploring flow and diffusion-based generative models with scalable interpolant transformers.
\newblock In \emph{European Conference on Computer Vision}. Springer, 2024{\natexlab{a}}.

\bibitem[Ma et~al.(2024{\natexlab{b}})Ma, Wang, Jia, Chen, Liu, Li, Chen, and Qiao]{ma2024latte}
Xin Ma, Yaohui Wang, Gengyun Jia, Xinyuan Chen, Ziwei Liu, Yuan-Fang Li, Cunjian Chen, and Yu Qiao.
\newblock Latte: Latent diffusion transformer for video generation.
\newblock \emph{arXiv preprint arXiv:2401.03048}, 2024{\natexlab{b}}.

\bibitem[Munoz et~al.(2021)Munoz, Zolfaghari, Argus, and Brox]{munoz2021temporal}
Andres Munoz, Mohammadreza Zolfaghari, Max Argus, and Thomas Brox.
\newblock Temporal shift {GAN} for large scale video generation.
\newblock In \emph{IEEE/CVF Winter Conference on Applications of Computer Vision}, 2021.

\bibitem[Peebles and Xie(2023)]{Peebles2022DiT}
William Peebles and Saining Xie.
\newblock Scalable diffusion models with transformers.
\newblock In \emph{IEEE International Conference on Computer Vision}, 2023.

\bibitem[Peng et~al.(2023)Peng, Alcaide, Anthony, Albalak, Arcadinho, Cao, Cheng, Chung, Grella, GV, et~al.]{peng2023rwkv}
Bo Peng, Eric Alcaide, Quentin Anthony, Alon Albalak, Samuel Arcadinho, Huanqi Cao, Xin Cheng, Michael Chung, Matteo Grella, Kranthi~Kiran GV, et~al.
\newblock Rwkv: Reinventing rnns for the transformer era.
\newblock \emph{arXiv preprint arXiv:2305.13048}, 2023.

\bibitem[Pope et~al.(2023)Pope, Douglas, Chowdhery, Devlin, Bradbury, Heek, Xiao, Agrawal, and Dean]{pope2023efficiently}
Reiner Pope, Sholto Douglas, Aakanksha Chowdhery, Jacob Devlin, James Bradbury, Jonathan Heek, Kefan Xiao, Shivani Agrawal, and Jeff Dean.
\newblock Efficiently scaling transformer inference.
\newblock \emph{Proceedings of Machine Learning and Systems}, 5, 2023.

\bibitem[Qiu et~al.(2024)Qiu, Xia, Zhang, He, Wang, Shan, and Liu]{qiu2023freenoise}
Haonan Qiu, Menghan Xia, Yong Zhang, Yingqing He, Xintao Wang, Ying Shan, and Ziwei Liu.
\newblock {FreeNoise}: Tuning-free longer video diffusion via noise rescheduling.
\newblock In \emph{International Conference on Learning Representations}, 2024.

\bibitem[Rakhimov et~al.(2020)Rakhimov, Volkhonskiy, Artemov, Zorin, and Burnaev]{rakhimov2020latent}
Ruslan Rakhimov, Denis Volkhonskiy, Alexey Artemov, Denis Zorin, and Evgeny Burnaev.
\newblock Latent video transformer.
\newblock \emph{arXiv preprint arXiv:2006.10704}, 2020.

\bibitem[Rombach et~al.(2022)Rombach, Blattmann, Lorenz, Esser, and Ommer]{rombach2021highresolution}
Robin Rombach, Andreas Blattmann, Dominik Lorenz, Patrick Esser, and Bj\"orn Ommer.
\newblock High-resolution image synthesis with latent diffusion models.
\newblock In \emph{IEEE Conference on Computer Vision and Pattern Recognition}, 2022.

\bibitem[Ruhe et~al.(2024)Ruhe, Heek, Salimans, and Hoogeboom]{ruhe2024rolling}
David Ruhe, Jonathan Heek, Tim Salimans, and Emiel Hoogeboom.
\newblock Rolling diffusion models.
\newblock \emph{arXiv preprint arXiv:2402.09470}, 2024.

\bibitem[Saito et~al.(2017)Saito, Matsumoto, and Saito]{saito2017temporal}
Masaki Saito, Eiichi Matsumoto, and Shunta Saito.
\newblock Temporal generative adversarial nets with singular value clipping.
\newblock In \emph{IEEE International Conference on Computer Vision}, 2017.

\bibitem[Seo et~al.(2022)Seo, Lee, Liu, James, and Abbeel]{seo2022autoregressive}
Younggyo Seo, Kimin Lee, Fangchen Liu, Stephen James, and Pieter Abbeel.
\newblock Autoregressive latent video prediction with high-fidelity image generator.
\newblock In \emph{IEEE International Conference on Image Processing}, 2022.

\bibitem[Singer et~al.(2023)Singer, Polyak, Hayes, Yin, An, Zhang, Hu, Yang, Ashual, Gafni, et~al.]{singer2022make}
Uriel Singer, Adam Polyak, Thomas Hayes, Xi Yin, Jie An, Songyang Zhang, Qiyuan Hu, Harry Yang, Oron Ashual, Oran Gafni, et~al.
\newblock Make-a-video: Text-to-video generation without text-video data.
\newblock In \emph{International Conference on Learning Representations}, 2023.

\bibitem[Sitzmann et~al.(2020)Sitzmann, Martel, Bergman, Lindell, and Wetzstein]{sitzmann2020implicit}
Vincent Sitzmann, Julien Martel, Alexander Bergman, David Lindell, and Gordon Wetzstein.
\newblock Implicit neural representations with periodic activation functions.
\newblock \emph{Advances in Neural Information Processing Systems}, 2020.

\bibitem[Skorokhodov et~al.(2021)Skorokhodov, Ignatyev, and Elhoseiny]{skorokhodov2021adversarial}
Ivan Skorokhodov, Savva Ignatyev, and Mohamed Elhoseiny.
\newblock Adversarial generation of continuous images.
\newblock In \emph{IEEE Conference on Computer Vision and Pattern Recognition}, 2021.

\bibitem[Skorokhodov et~al.(2022)Skorokhodov, Tulyakov, and Elhoseiny]{skorokhodov2021stylegan}
Ivan Skorokhodov, Sergey Tulyakov, and Mohamed Elhoseiny.
\newblock {StyleGAN-V}: A continuous video generator with the price, image quality and perks of {StyleGAN2}.
\newblock In \emph{IEEE Conference on Computer Vision and Pattern Recognition}, 2022.

\bibitem[Song et~al.(2021{\natexlab{a}})Song, Meng, and Ermon]{song2021denoising}
Jiaming Song, Chenlin Meng, and Stefano Ermon.
\newblock Denoising diffusion implicit models.
\newblock In \emph{International Conference on Learning Representations}, 2021{\natexlab{a}}.

\bibitem[Song and Ermon(2019)]{song2019generative}
Yang Song and Stefano Ermon.
\newblock Generative modeling by estimating gradients of the data distribution.
\newblock In \emph{Advances in Neural Information Processing Systems}, 2019.

\bibitem[Song et~al.(2021{\natexlab{b}})Song, Sohl-Dickstein, Kingma, Kumar, Ermon, and Poole]{song2021scorebased}
Yang Song, Jascha Sohl-Dickstein, Diederik~P Kingma, Abhishek Kumar, Stefano Ermon, and Ben Poole.
\newblock Score-based generative modeling through stochastic differential equations.
\newblock In \emph{International Conference on Learning Representations}, 2021{\natexlab{b}}.

\bibitem[Soomro et~al.(2012)Soomro, Zamir, and Shah]{soomro2012ucf101}
Khurram Soomro, Amir~Roshan Zamir, and Mubarak Shah.
\newblock {UCF101}: A dataset of 101 human actions classes from videos in the wild.
\newblock \emph{arXiv preprint arXiv:1212.0402}, 2012.

\bibitem[Srivastava et~al.(2017)Srivastava, Valkov, Russell, Gutmann, and Sutton]{srivastava2017veegan}
Akash Srivastava, Lazar Valkov, Chris Russell, Michael~U Gutmann, and Charles Sutton.
\newblock Veegan: Reducing mode collapse in gans using implicit variational learning.
\newblock \emph{Advances in Neural Information Processing Systems}, 2017.

\bibitem[Srivastava et~al.(2015)Srivastava, Mansimov, and Salakhudinov]{srivastava2015unsupervised}
Nitish Srivastava, Elman Mansimov, and Ruslan Salakhudinov.
\newblock Unsupervised learning of video representations using {LSTM}s.
\newblock In \emph{International Conference on Machine Learning}, 2015.

\bibitem[Su et~al.(2024)Su, Ahmed, Lu, Pan, Bo, and Liu]{su2024roformer}
Jianlin Su, Murtadha Ahmed, Yu Lu, Shengfeng Pan, Wen Bo, and Yunfeng Liu.
\newblock Roformer: Enhanced transformer with rotary position embedding.
\newblock \emph{Neurocomputing}, 568:\penalty0 127063, 2024.

\bibitem[Tian et~al.(2021)Tian, Ren, Chai, Olszewski, Peng, Metaxas, and Tulyakov]{tian2021good}
Yu Tian, Jian Ren, Menglei Chai, Kyle Olszewski, Xi Peng, Dimitris~N Metaxas, and Sergey Tulyakov.
\newblock A good image generator is what you need for high-resolution video synthesis.
\newblock In \emph{International Conference on Learning Representations}, 2021.

\bibitem[Tran et~al.(2015)Tran, Bourdev, Fergus, Torresani, and Paluri]{tran2015learning}
Du Tran, Lubomir Bourdev, Rob Fergus, Lorenzo Torresani, and Manohar Paluri.
\newblock Learning spatiotemporal features with 3{D} convolutional networks.
\newblock In \emph{IEEE International Conference on Computer Vision}, 2015.

\bibitem[Tulyakov et~al.(2018)Tulyakov, Liu, Yang, and Kautz]{tulyakov2018mocogan}
Sergey Tulyakov, Ming-Yu Liu, Xiaodong Yang, and Jan Kautz.
\newblock {MoCoGAN}: Decomposing motion and content for video generation.
\newblock In \emph{IEEE Conference on Computer Vision and Pattern Recognition}, 2018.

\bibitem[Unterthiner et~al.(2018)Unterthiner, van Steenkiste, Kurach, Marinier, Michalski, and Gelly]{unterthiner2018towards}
Thomas Unterthiner, Sjoerd van Steenkiste, Karol Kurach, Raphael Marinier, Marcin Michalski, and Sylvain Gelly.
\newblock Towards accurate generative models of video: A new metric \& challenges.
\newblock \emph{arXiv preprint arXiv:1812.01717}, 2018.

\bibitem[van~den Oord et~al.(2017)van~den Oord, Vinyals, and Kavukcuoglu]{van2017neural}
Aaron van~den Oord, Oriol Vinyals, and Koray Kavukcuoglu.
\newblock Neural discrete representation learning.
\newblock In \emph{Advances in Neural Information Processing Systems}, 2017.

\bibitem[Vaswani et~al.(2017)Vaswani, Shazeer, Parmar, Uszkoreit, Jones, Gomez, Kaiser, and Polosukhin]{vaswani2017attention}
Ashish Vaswani, Noam Shazeer, Niki Parmar, Jakob Uszkoreit, Llion Jones, Aidan~N Gomez, {\L}ukasz Kaiser, and Illia Polosukhin.
\newblock Attention is all you need.
\newblock In \emph{Advances in Neural Information Processing Systems}, 2017.

\bibitem[Villegas et~al.(2019)Villegas, Pathak, Kannan, Erhan, Le, and Lee]{villegas2019high}
Ruben Villegas, Arkanath Pathak, Harini Kannan, Dumitru Erhan, Quoc~V Le, and Honglak Lee.
\newblock High fidelity video prediction with large stochastic recurrent neural networks.
\newblock In \emph{Advances in Neural Information Processing Systems}, 2019.

\bibitem[Villegas et~al.(2023)Villegas, Babaeizadeh, Kindermans, Moraldo, Zhang, Saffar, Castro, Kunze, and Erhan]{villegas2023phenaki}
Ruben Villegas, Mohammad Babaeizadeh, Pieter-Jan Kindermans, Hernan Moraldo, Han Zhang, Mohammad~Taghi Saffar, Santiago Castro, Julius Kunze, and Dumitru Erhan.
\newblock Phenaki: Variable length video generation from open domain textual descriptions.
\newblock In \emph{International Conference on Learning Representations}, 2023.

\bibitem[Voleti et~al.(2022)Voleti, Jolicoeur-Martineau, and Pal]{voleti2022mcvd}
Vikram Voleti, Alexia Jolicoeur-Martineau, and Chris Pal.
\newblock Mcvd-masked conditional video diffusion for prediction, generation, and interpolation.
\newblock \emph{Advances in Neural Information Processing Systems}, 2022.

\bibitem[Vondrick et~al.(2016)Vondrick, Pirsiavash, and Torralba]{vondrick2016generating}
Carl Vondrick, Hamed Pirsiavash, and Antonio Torralba.
\newblock Generating videos with scene dynamics.
\newblock In \emph{Advances in Neural Information Processing Systems}, 2016.

\bibitem[Wang et~al.(2020)Wang, Li, Khabsa, Fang, and Ma]{wang2020linformer}
Sinong Wang, Belinda~Z Li, Madian Khabsa, Han Fang, and Hao Ma.
\newblock Linformer: Self-attention with linear complexity.
\newblock \emph{arXiv preprint arXiv:2006.04768}, 2020.

\bibitem[Wang et~al.(2023)Wang, Yang, Tuo, He, Zhu, Fu, and Liu]{wang2023videofactory}
Wenjing Wang, Huan Yang, Zixi Tuo, Huiguo He, Junchen Zhu, Jianlong Fu, and Jiaying Liu.
\newblock Videofactory: Swap attention in spatiotemporal diffusions for text-to-video generation.
\newblock \emph{arXiv preprint arXiv:2305.10874}, 2023.

\bibitem[Weissenborn et~al.(2020)Weissenborn, T{\"a}ckstr{\"o}m, and Uszkoreit]{weissenborn2020scaling}
Dirk Weissenborn, Oscar T{\"a}ckstr{\"o}m, and Jakob Uszkoreit.
\newblock Scaling autoregressive video models.
\newblock In \emph{International Conference on Learning Representations}, 2020.

\bibitem[Weng et~al.(2023)Weng, Feng, Wang, Dai, Wang, Yin, Zhao, Qiu, Bao, Yuan, Luo, Zhang, and Xiong]{weng2023art}
Wenming Weng, Ruoyu Feng, Yanhui Wang, Qi Dai, Chunyu Wang, Dacheng Yin, Zhiyuan Zhao, Kai Qiu, Jianmin Bao, Yuhui Yuan, Chong Luo, Yueyi Zhang, and Zhiwei Xiong.
\newblock Art•v: Auto-regressive text-to-video generation with diffusion models.
\newblock \emph{arXiv preprint arXiv:2311.18834}, 2023.

\bibitem[Wu et~al.(2022)Wu, Rabe, Hutchins, and Szegedy]{wu2022memorizing}
Yuhuai Wu, Markus~N Rabe, DeLesley Hutchins, and Christian Szegedy.
\newblock Memorizing transformers.
\newblock \emph{arXiv preprint arXiv:2203.08913}, 2022.

\bibitem[Xie et~al.(2024)Xie, Xu, Hong, Tan, Liu, Liu, Kaufman, and Zhou]{xie2024progressive}
Desai Xie, Zhan Xu, Yicong Hong, Hao Tan, Difan Liu, Feng Liu, Arie Kaufman, and Yang Zhou.
\newblock Progressive autoregressive video diffusion models.
\newblock \emph{arXiv preprint arXiv:2410.08151}, 2024.

\bibitem[Yan et~al.(2021)Yan, Zhang, Abbeel, and Srinivas]{yan2021videogpt}
Wilson Yan, Yunzhi Zhang, Pieter Abbeel, and Aravind Srinivas.
\newblock {VideoGPT}: Video generation using {VQ-VAE} and transformers.
\newblock \emph{arXiv preprint arXiv:2104.10157}, 2021.

\bibitem[Yan et~al.(2023)Yan, Hafner, James, and Abbeel]{yan2023temporally}
Wilson Yan, Danijar Hafner, Stephen James, and Pieter Abbeel.
\newblock Temporally consistent transformers for video generation.
\newblock In \emph{International Conference on Machine Learning}, 2023.

\bibitem[Yang et~al.(2022)Yang, Srivastava, and Mandt]{yang2022diffusion}
Ruihan Yang, Prakhar Srivastava, and Stephan Mandt.
\newblock Diffusion probabilistic modeling for video generation.
\newblock \emph{arXiv preprint arXiv:2203.09481}, 2022.

\bibitem[Yin et~al.(2023)Yin, Wu, Yang, Wang, Wang, Ni, Yang, Li, Liu, Yang, Fu, Gong, Wang, Liu, Li, and Duan]{yin2023nuwa}
Shengming Yin, Chenfei Wu, Huan Yang, Jianfeng Wang, Xiaodong Wang, Minheng Ni, Zhengyuan Yang, Linjie Li, Shuguang Liu, Fan Yang, Jianlong Fu, Ming Gong, Lijuan Wang, Zicheng Liu, Houqiang Li, and Nan Duan.
\newblock {NUWA}-{XL}: Diffusion over diffusion for e{X}tremely long video generation.
\newblock In \emph{Proceedings of the 61st Annual Meeting of the Association for Computational Linguistics (Volume 1: Long Papers)}, 2023.

\bibitem[Yoo et~al.(2023)Yoo, Kim, Lee, Kim, and Hong]{yoo2023towards}
Jaehoon Yoo, Semin Kim, Doyup Lee, Chiheon Kim, and Seunghoon Hong.
\newblock Towards end-to-end generative modeling of long videos with memory-efficient bidirectional transformers.
\newblock In \emph{IEEE Conference on Computer Vision and Pattern Recognition}, 2023.

\bibitem[Yu et~al.(2022{\natexlab{a}})Yu, Li, Koh, Zhang, Pang, Qin, Ku, Xu, Baldridge, and Wu]{yu2022vectorquantized}
Jiahui Yu, Xin Li, Jing~Yu Koh, Han Zhang, Ruoming Pang, James Qin, Alexander Ku, Yuanzhong Xu, Jason Baldridge, and Yonghui Wu.
\newblock Vector-quantized image modeling with improved {VQGAN}.
\newblock In \emph{International Conference on Learning Representations}, 2022{\natexlab{a}}.

\bibitem[Yu et~al.(2023{\natexlab{a}})Yu, Cheng, Sohn, Lezama, Zhang, Chang, Hauptmann, Yang, Hao, Essa, et~al.]{yu2023magvit}
Lijun Yu, Yong Cheng, Kihyuk Sohn, Jos{\'e} Lezama, Han Zhang, Huiwen Chang, Alexander~G Hauptmann, Ming-Hsuan Yang, Yuan Hao, Irfan Essa, et~al.
\newblock Magvit: Masked generative video transformer.
\newblock In \emph{IEEE Conference on Computer Vision and Pattern Recognition}, 2023{\natexlab{a}}.

\bibitem[Yu et~al.(2024{\natexlab{a}})Yu, Lezama, Gundavarapu, Versari, Sohn, Minnen, Cheng, Gupta, Gu, Hauptmann, Gong, Yang, Essa, Ross, and Jiang]{yu2024language}
Lijun Yu, Jose Lezama, Nitesh~Bharadwaj Gundavarapu, Luca Versari, Kihyuk Sohn, David Minnen, Yong Cheng, Agrim Gupta, Xiuye Gu, Alexander~G Hauptmann, Boqing Gong, Ming-Hsuan Yang, Irfan Essa, David~A Ross, and Lu Jiang.
\newblock Language model beats diffusion - tokenizer is key to visual generation.
\newblock In \emph{International Conference on Learning Representations}, 2024{\natexlab{a}}.

\bibitem[Yu et~al.(2022{\natexlab{b}})Yu, Tack, Mo, Kim, Kim, Ha, and Shin]{yu2022digan}
Sihyun Yu, Jihoon Tack, Sangwoo Mo, Hyunsu Kim, Junho Kim, Jung-Woo Ha, and Jinwoo Shin.
\newblock Generating videos with dynamics-aware implicit generative adversarial networks.
\newblock In \emph{International Conference on Learning Representations}, 2022{\natexlab{b}}.

\bibitem[Yu et~al.(2023{\natexlab{b}})Yu, Sohn, Kim, and Shin]{yu2023video}
Sihyun Yu, Kihyuk Sohn, Subin Kim, and Jinwoo Shin.
\newblock Video probabilistic diffusion models in projected latent space.
\newblock In \emph{IEEE Conference on Computer Vision and Pattern Recognition}, 2023{\natexlab{b}}.

\bibitem[Yu et~al.(2024{\natexlab{b}})Yu, Kwak, Jang, Jeong, Huang, Shin, and Xie]{yu2024representation}
Sihyun Yu, Sangkyung Kwak, Huiwon Jang, Jongheon Jeong, Jonathan Huang, Jinwoo Shin, and Saining Xie.
\newblock Representation alignment for generation: Training diffusion transformers is easier than you think.
\newblock \emph{arXiv preprint arXiv:2410.06940}, 2024{\natexlab{b}}.

\bibitem[Yu et~al.(2024{\natexlab{c}})Yu, Nie, Huang, Li, Shin, and Anandkumar]{yu2024efficient}
Sihyun Yu, Weili Nie, De-An Huang, Boyi Li, Jinwoo Shin, and Anima Anandkumar.
\newblock Efficient video diffusion models via content-frame motion-latent decomposition.
\newblock In \emph{International Conference on Learning Representations}, 2024{\natexlab{c}}.

\bibitem[Yushchenko et~al.(2019)Yushchenko, Araslanov, and Roth]{yushchenko2019markov}
Vladyslav Yushchenko, Nikita Araslanov, and Stefan Roth.
\newblock Markov decision process for video generation.
\newblock In \emph{Proceedings of the IEEE/CVF International Conference on Computer Vision Workshops}, 2019.

\bibitem[Zeng et~al.(2022)Zeng, Vahdat, Williams, Gojcic, Litany, Fidler, and Kreis]{zeng2022lion}
Xiaohui Zeng, Arash Vahdat, Francis Williams, Zan Gojcic, Or Litany, Sanja Fidler, and Karsten Kreis.
\newblock {LION}: Latent point diffusion models for 3d shape generation.
\newblock In \emph{Advances in Neural Information Processing Systems}, 2022.

\bibitem[Zhang et~al.(2018)Zhang, Isola, Efros, Shechtman, and Wang]{zhang2018perceptual}
Richard Zhang, Phillip Isola, Alexei~A Efros, Eli Shechtman, and Oliver Wang.
\newblock The unreasonable effectiveness of deep features as a perceptual metric.
\newblock In \emph{IEEE Conference on Computer Vision and Pattern Recognition}, 2018.

\bibitem[Zheng et~al.(2024)Zheng, Nie, Vahdat, and Anandkumar]{zheng2024fast}
Hongkai Zheng, Weili Nie, Arash Vahdat, and Anima Anandkumar.
\newblock Fast training of diffusion models with masked transformers.
\newblock \emph{Transactions on Machine Learning Research}, 2024.

\bibitem[Zhou et~al.(2022)Zhou, Wang, Yan, Lv, Zhu, and Feng]{zhou2022magicvideo}
Daquan Zhou, Weimin Wang, Hanshu Yan, Weiwei Lv, Yizhe Zhu, and Jiashi Feng.
\newblock Magicvideo: Efficient video generation with latent diffusion models.
\newblock \emph{arXiv preprint arXiv:2211.11018}, 2022.

\end{thebibliography}
}

\appendix
\maketitlesupplementary
\setcounter{section}{0}  
\section{Sampling Procedure}
\label{appen:sampling}
We provide detailed sampling procedure of \sname in Algorithm~\ref{algo:sampling}.

\begin{algorithm}[h!]
\begin{spacing}{1.05}
\caption{\sname Diffusion}\label{algo:sampling}
\begin{algorithmic}[1]
\For{$n=1$ to $N$} \Comment{{\it Autoregressively generate $n$-th latent vector $\bz^n$.}}
\State Sample the random noise $\bz_T^{n} \sim p(\bz_T)$.
\For{$i$ in $\{0, \ldots, M-1\}$}
\State Compute the score $\bm{\epsilon}_i \leftarrow D_{\bm{\theta}}(\bz_{i}^{n}, t;\, \bh^{n-1}, \bc)$.
\State Compute $\bz_{i+1}^n \leftarrow \bz_i^n + (t_{i+1} - t_{i})\bm{\epsilon}_i$ \Comment{{\it Euler solver; can be different with other solvers.}}
\EndFor
\State $\bh^{n} = \mathrm{HiddenState}\big(D_{\bm{\theta}} (\bz_M^{n}, 0;\, \bh^{i-1}, \bc)\big)$ \Comment{{\it Compute memory latent vector.}}
\EndFor
\State Decode $[\bx^1,\ldots,\bx^{N}]$ from generated latent vectors $[\bz^1,\ldots,\bz^{N}]$.
\State Output the generated video $[\bx^1,\ldots,\bx^{N}]$.
\end{algorithmic}
\end{spacing}
\end{algorithm}

\section{Architecture Illustration}
\label{appen:archi}

Figure~\ref{fig:detailed_architecture} visualizes the \sname architecture in detail as applied to the existing W.A.L.T~\citep{gupta2023photorealistic} architecture. W.A.L.T is a variant of diffusion transformer (DiT)~\citep{Peebles2022DiT} that uses repeating spatiotemporal and spatial window attention layers instead of full attention. As illustrated in this figure, \sname adds memory attention layers at the beginning of each spatiotemporal window attention layer and spatial window attention layer. Here, window attention layers are similar to the original DiT that uses adaptive instance normalization (AdaIN), where we use LoRA~\cite{hu2021lora} for parametrization of MLP proposed by W.A.L.T to reduce the number of parameters without performance degradation. For memory attention layer, we simply add a cross attention layer without AdaIN.
\section{More Discussion with Related Work}
\label{appen:related}

\textbf{Diffusion model for sequential data.}
We emphasize that although we primarily explore video data in this paper, our method is not limited to video data. There exists multiple works that try to generate general sequential data (\eg, climate data, PDE data, audio, \emph{etc}.) through diffusion models. For instance, \citet{ruhe2024rolling} formulates a diffusion process specialized for sequential data; they propose denoising process with a \emph{sliding window}, where the noise scale of elements within a window is set differently by considering uncertainty of each element. Another line of work~\citep{ge2023preserve,lu2024improve} have tried to solve this problem by exploring temporal correlation of Gaussian noise in diffusion process with \emph{fixed-size} sequences. \citet{lippe2023pderefiner} tries to solve PDE through diffusion model based on autoregressive rollout, and demonstrates high-frequency details can be predicted better than existing PDE solvers. Our method primarily focuses on generating video data, where each element (\ie, video frame) is already quite high-dimensional and thus mitigating error propagation and ensuring large context window size is more difficult than other temporal data. Therefore, we believe our method can be extending to other forms of sequential data listed above. 

\vspace{0.02in}
\noindent\textbf{Video generation.}
There are many works to tackle the problem of video generation. First, there are approaches that uses generative adversarial network (GAN;~\citet{goodfellow2014generative}) for modeling video distribution \citep{tulyakov2018mocogan,yu2022digan,skorokhodov2021stylegan,tian2021good,acharya2018towards,clark2019adversarial,fox2021stylevideogan,gordon2021latent,kahembwe2020lower,munoz2021temporal,saito2017temporal,vondrick2016generating,yushchenko2019markov}. They usually extend popular image GAN architectures (\eg, StyleGAN~\citep{karras2020analyzing}) by considering an additional temporal axis. Another line of works encode videos in discrete token space using popular vector-quantized autoencoders~\citep{yu2024language,van2017neural,yu2022vectorquantized} and learn video distribution in this latent space, either with autoregressive transformers \citep{yu2024language,ge2022long,yan2021videogpt,kalchbrenner2017video,rakhimov2020latent,weissenborn2020scaling} or masked generative transformers \citep{yoo2023towards,yu2023magvit}. Finally, many of recent works propose diffusion-based approach for generating videos~\citep{ho2022video,harvey2022flexible,lu2023vdt,singer2022make,weng2023art,hoppe2022diffusion,yang2022diffusion}, and some of them have attempt to exploit knowledge learned from large-scale image datasets \citep{blattmann2023align,he2022lvdm,ge2023preserve,ho2022imagen,singer2022make,wang2023videofactory,an2023latent} by fine-tuning image diffusion models or do image-video joint training. Our model is also categorized as diffusion-based approach to synthesize long videos.

\vspace{0.02in}
\noindent\textbf{Video prediction.}
As our model has a capability of both video generation and prediction, it also has a close relationship to existing video prediction methods \citep{srivastava2015unsupervised,finn2016unsupervised,denton2017unsupervised,babaeizadeh2018stochastic,denton2018stochastic,lee2018stochastic,villegas2019high,kumar2020videoflow,franceschi2020stochastic,luc2020transformation,lee2021revisiting,seo2022autoregressive}. These methods usually use given frames as condition to the model and generate future frames via popular deep generative models, such as GANs, diffusion models, and autoregressive  long video prediction~\citep{harvey2022flexible,yan2023temporally}, in contrast to previous works that usually focus on predicting a few video frames. Our method also shares a similar goal to predict video frames with a long horizon.

\twocolumn[{
\centering
    \includegraphics[width=.84\linewidth]{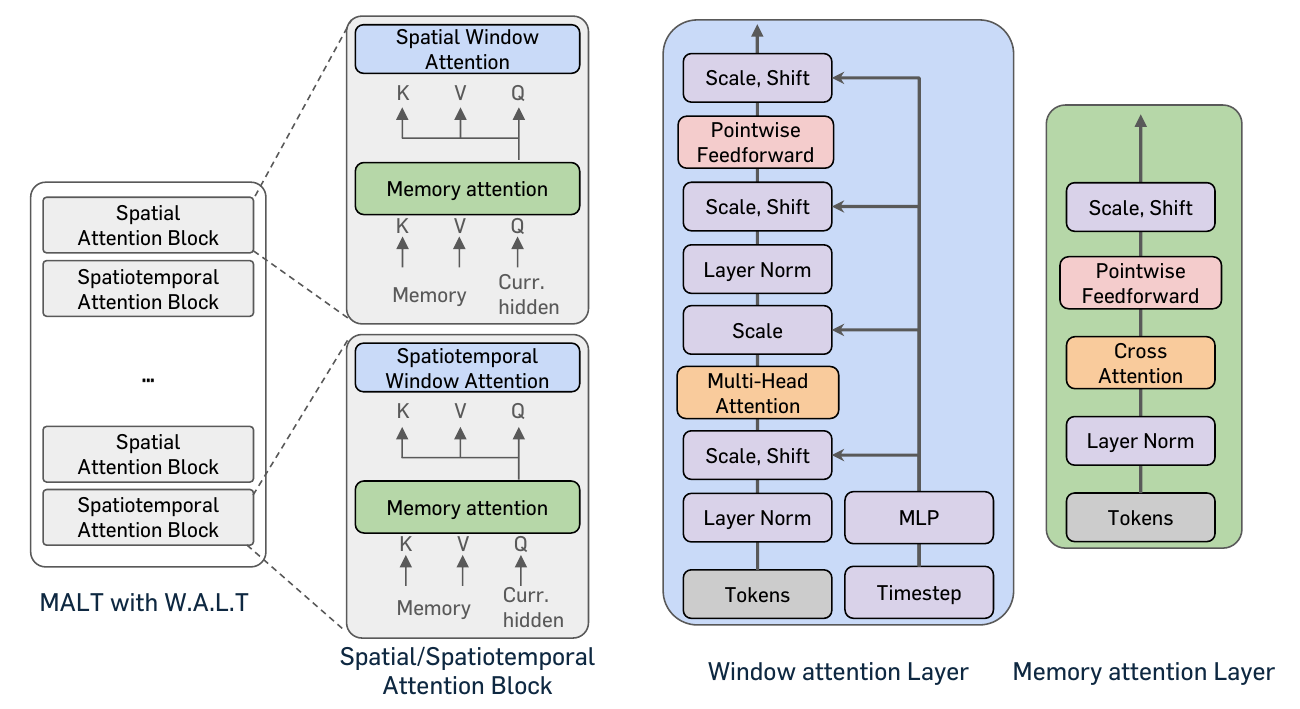}
    \captionof{figure}{\textbf{Detailed architecture illustration} of \sname if it is incoporated with W.A.L.T~\citep{gupta2023photorealistic}. 
    } 
    \label{fig:detailed_architecture}
    \vspace{0.2in}
}]

\section{Setup}
\label{appen:setup}

\subsection{Datasets}
\label{appen:dataset}
\textbf{UCF-101.}
UCF-101 \citep{soomro2012ucf101} is a video dataset widely used for evaluation of video generation methods. Is consists of 101 classes of different human actions, where each video has 320$\times$240 resolution frames. The dataset includes 13,320 videos with 9,537 training split and 3,783 test split. Following the previous recent video generation literature we only use train split and test split for evaluation~\citep{yu2023magvit, singer2022make}. We center-cropped each video and resized it into $128\times128$ resolution. We use first 128 frames of videos for training and evaluation.

\vspace{0.02in}
\noindent\textbf{Kinetics-600.} Kinetics-600~\citep{kay2017kinetics} is a large-scale complex video dataset consisting of 600 action categories. It contains about 480,000 videos in total, where they are divided into 390,000, 30,000, and 60,000 videos for train, validation, and test splits (respectively). Following the previous video prediction benchmarks~\citep{gupta2023photorealistic,yan2023temporally,yu2023magvit}, we center-crop each video frame and resize each video frame as 128$\times$128 resolution. Due to the different tokenizer used in our method, we use 17 frames for condition (a bit different from TECO~\citep{yan2023temporally} that uses 20 frames for condition) and but predict 80 frames same as TECO. We use train split for train and validation split for evaluation.

\vspace{0.02in}
\textbf{T2V dataset.}
We use 89M text-short-video pairs (up to 37 frames) and 970M text-image pairs from public and internal sources. All datasets are center-cropped and resized to 128$\times$128 resolution. For videos, we sample the frames with a fps of 8.

\subsection{Hyperparameters}
We mostly follow the same hyperparameter setup to W.A.L.T~\citep{gupta2023photorealistic}. Specifically, we provide the detailed hyperparameter values in Table~\ref{tab:hyperparam}.

\begin{table}[!ht]
    \caption{Hyperparameter setup.}
    \centering
    \resizebox{!}{.6\linewidth}{%
    \begin{tabular}{l c c c}
        \toprule
         & {UCF-101} & {Kinetics-600} & T2V  \\
        \midrule
        \textbf{Autoencoder} \\
        Input dim. & 17$\times$128$\times$128$\times$3 & 17$\times$128$\times$128$\times$3 & 17$\times$128$\times$128$\times$3 \\
        $d_s$, $d_t$ & {8, 4} & {8, 4} & {8, 4} \\
        Channel dim. & 128 & 128 & 128 \\
        Channel multiplier & {1, 2, 2, 4} & {1, 2, 2, 4} & {1, 2, 2, 4} \\
        Training duration & 2,000 epochs & 270,000 steps & 1,000,000 steps \\ 
        Batch size & 256 & 256 & 256 \\
        lr scheduler & Cosine & Cosine & Cosine \\
        Optimizer & Adam & Adam & Adam \\
        \midrule
        \textbf{Diffusion model} \\
        Input dim. & 5$\times$16$\times$16 & 5$\times$16$\times$16 & 5$\times$16$\times$16 \\
        Num. layers & 28 & 24 & 24 \\
        Hidden dim. & 1,152 & 1,024 & 1,024 \\
        Num. heads & 16 & 16 & 16 \\ 
        $\sigma_{\mathrm{mem}}$ & 0.1 & 0.1 & 0.1 \\
        $N$ & 7 & 5 & 2 \\
        Training duration & 120,000 steps & 270,000 steps & 700,000 steps \\ 
        Batch size & 256 & 256 & 256 \\ 
        lr scheduler & Cosine & Cosine & Cosine \\
        Optimizer & AdamW & AdamW & AdamW \\
        lr & 0.0005 & 0.0005 &  0.0002 \\
        \midrule
        \textbf{Diffusion} \\
        Diffusion steps & 1000 & 1000 & 1000 \\
        Noise schedule & Linear & Linear & Linear  \\
        $\beta_0$ & 0.0001 & 0.0001 & 0.0001 \\
        $\beta_T$ & 0.02 & 0.02 & 0.02 \\
        Training objective & v-prediction & v-prediction & v-prediction \\
        Sampler & DDIM~\citep{song2021denoising} & DDIM~\citep{song2021denoising} & DDIM~\citep{song2021denoising} \\
        Sampling steps & 50 & 50 & 50 \\
        Guidance & - & - & \checkmark \\
        \bottomrule
         & 
    \end{tabular}
    }
    \label{tab:hyperparam}
    \vspace{-0.2in}
\end{table}

\subsection{Metrics}
\label{appen:metrics}
For Fr\'echet video distance (FVD; \citet{unterthiner2018towards}), we use the same protocol used in previous popular video generation works~\citep{yu2022digan, tulyakov2018mocogan}; we use a pretrained I3D network~\citep{tran2015learning} to compute feature for evaluating statistics to compute FVD. For UCF-101, we use 2,048 real and fame samples following the common practice in previous video generation literature~\citep{yu2022digan, skorokhodov2021adversarial,yu2023video}. For Kinetics-600, we use 256 samples for FVD evaluation and average the values of 4 runs, following the setup in TECO~\citep{yan2023temporally}. For other metrics (PSNR, SSIM, LPIPS) used in evaluation on Kinetics-600, we also follow the evaluation setup of TECO: we compute them in per-frame manner between predicted frames and ground-truth frames, and then average them.

\subsection{Training resources}
All experiments are conducted with Google Cloud TPU v5e 16$\times$16 instances, where each chip has a 16GB HBM2 capacity. Note that training can be done on devices with relative small memory (less than 16GB) because of our efficient training scheme design.

\label{appen:hyper}

\section{Baselines}
\label{appen:baselines}
In what follows, we explain the main idea of baseline methods that we used for the evaluation.
\begin{itemize}[leftmargin=0.2in]
\item \textbf{MoCoGAN}~\citep{tulyakov2018mocogan} proposes a video GAN to generate videos by decomposing motion and content of videos into two different latent vectors.
\item \textbf{MoCoGAN-HD}~\citep{tian2021good} also proposes a video GAN based on motion-content decomposition but uses a latent space of pretrained image GAN to achieve the goal.
\item \textbf{DIGAN}~\citep{yu2022digan} proposes to represent videos as implicit neural representations (INRs)~\citep{sitzmann2020implicit} and introduces a GAN to generate such INR parameters.
\item \textbf{StyleGAN-V}~\citep{skorokhodov2021stylegan} interprets videos as continuous function of time $t$ and extend StyleGAN-2~\citep{karras2020analyzing} architecture to efficiently learn long video distribution.
\item \textbf{PVDM}~\citep{yu2023video} proposes a latent diffusion model based on triplane-based encoding of videos~\citep{kim2022scalable} to avoid usage of computational-heavy 3D convolutions.
\item \textbf{HVDM}~\citep{kim2024hybrid} uses the ideas in PVDM but also incorporates 3D wavelet representation for video encoding to achieve better video reconstruction and generation.
\item \textbf{(Latent) FDM}~\citep{harvey2022flexible} proposes a diffusion model framework for long videos, by exploring various schemes to choose frames to be noised in the target long video tensor.
\item \textbf{Perceiver AR}~\citep{hawthorne2022general} proposes an autoregressive model that can handle long contexts in efficient and in domain-agnostic manner.
\item \textbf{CoordTok}~\citep{jang2024efficient} presents a continuous video tokenizer that can encode long videos into compact triplane latent representations. They also show this latent representations greatly improve generation efficiency and efficacy.
\item \textbf{TECO}~\citep{yan2023temporally} proposes a masked generative transformer~\citep{chang2022maskgit} specialized for long video generation, based on additional compression of image latent vector from image VQGAN~\citep{yu2022vectorquantized} and causal transformer to encode these compressed sequences.
\end{itemize}
\section{Additional Analysis}
\label{appen:ablation}
\begin{table}[ht!]
    \captionof{table}{\textbf{Additional analysis.} FVD, PSNR, and SSIM values on Kinetics-600. Gray indicates that the component is fixed for ablation of other components.}
    \centering\small
    \begin{tabular}{l c c c c c c c c}
    \toprule
    & {} &  {}  & FVD  & PSNR & SSIM\\
    \midrule
    \multirow{2}{*}{$\sigma_{\mathrm{mem}}$}  
    & {0.2} & {\textcolor{lightgray}{97}} & 382 & 15.0 & 0.426   \\
    & {0.1} & {\textcolor{lightgray}{97}} & 392 & 15.4 & 0.437 \\
    \midrule
    \multirow{2}{*}{\shortstack[l]{Training \\ Length}}
    & {\textcolor{lightgray}{0.1}}  & {57} & 437 & 14.9 & 0.421 \\
    & {\textcolor{lightgray}{0.1}}  & {97} & 392 & 15.4 & 0.437  \\
    \bottomrule
    \end{tabular}
    \label{tab:addi_abla}
\end{table}
In addition to experiments in Section~\ref{subsec:ablation}, we perform several additional experiments to show the effect of hyperparameter $\sigma_{\mathrm{mem}}$ and the length of video (\ie, number of segments) used in training, and report these results in Table~\ref{tab:addi_abla}. As shown in this table, \sname is quite robust the value choice of $\sigma_{\mathrm{mem}}$, and training \sname on longer videos improves the model to generate longer context.

\section{Additional Qualitative Results}
\label{appen:more_qual}
In what follows, we provide video generation results from MALT, mostly on long text-to-video generation results.
\begin{figure*}[ht!]
    \centering
    \includegraphics[width=.8\textwidth]{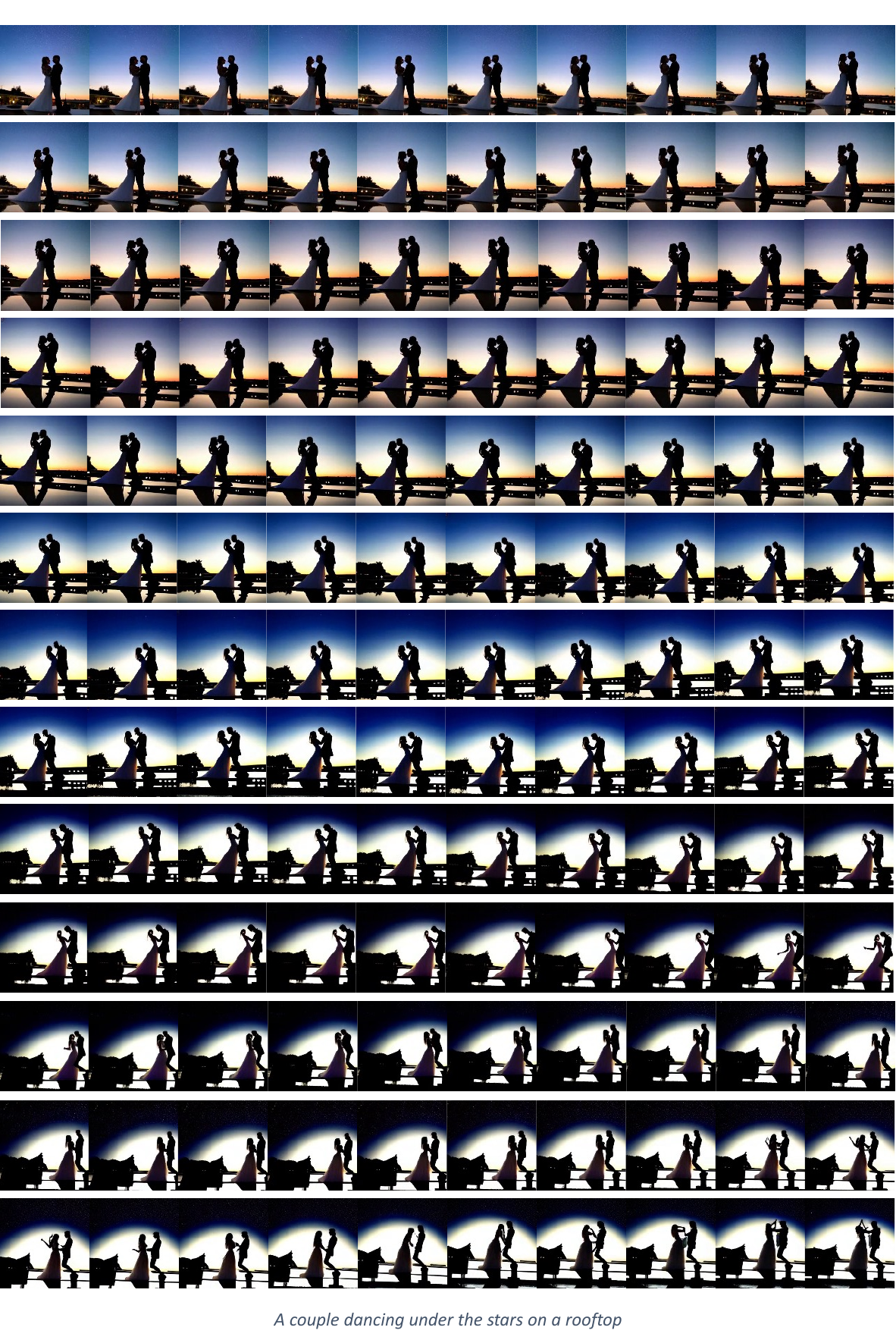}
    \caption{\textbf{Long text-to-video generation results} from \sname. We visualize video frames with a stride of 5. We visualize first 650 frames here and the next 600 frames are visualized in Figure~\ref{fig:t2v_supp_4}.
    } 
    \label{fig:t2v_supp_3}
\end{figure*}
\begin{figure*}[ht!]
    \centering
    \includegraphics[width=.8\textwidth]{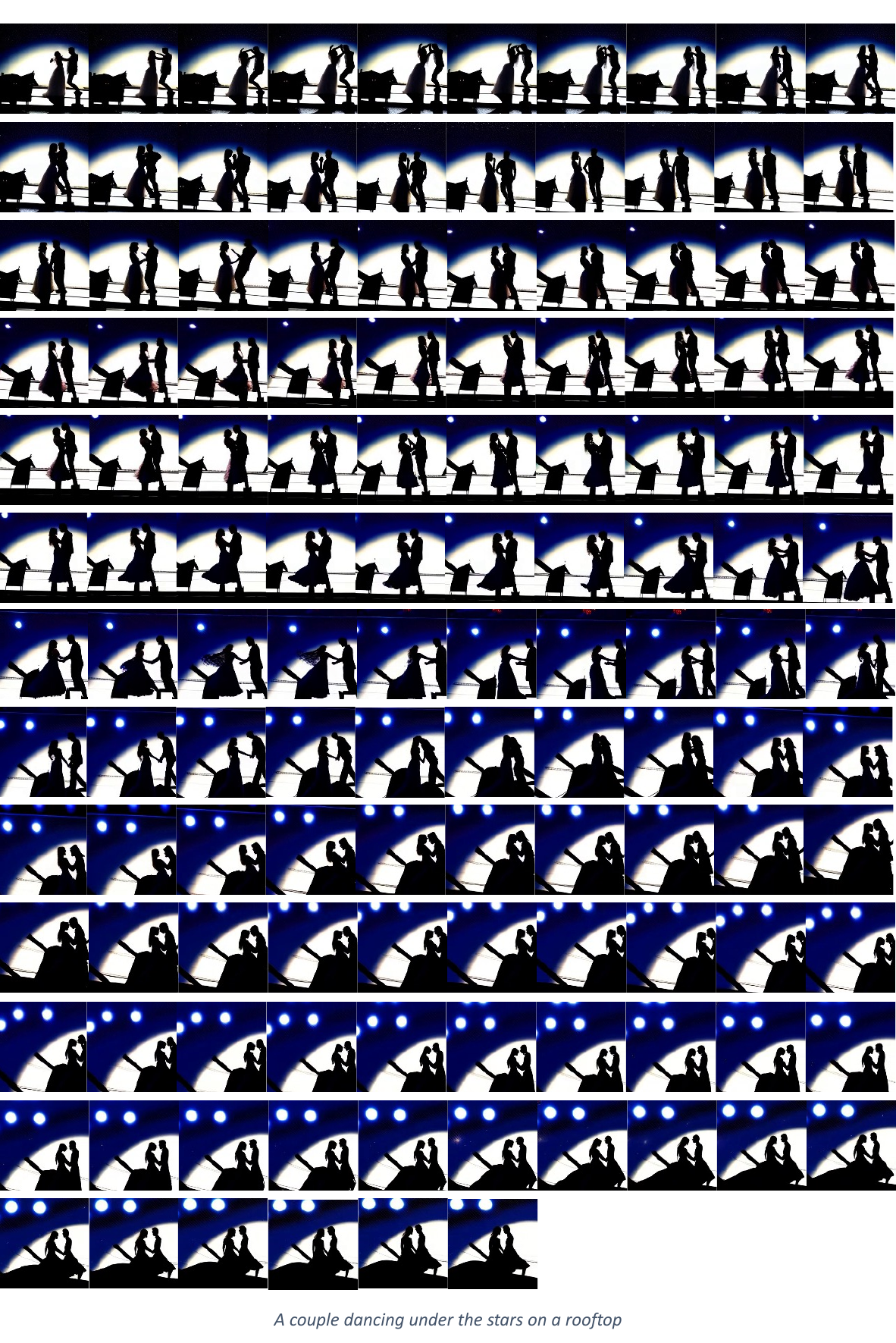}
    \caption{\textbf{Long text-to-video generation results} from \sname. We visualize video frames with a stride of 5. Video frames are continued from Figure~\ref{fig:t2v_supp_3}.
    } 
    \label{fig:t2v_supp_4}
\end{figure*}
\begin{figure*}[ht!]
    \centering
    \includegraphics[width=.8\textwidth]{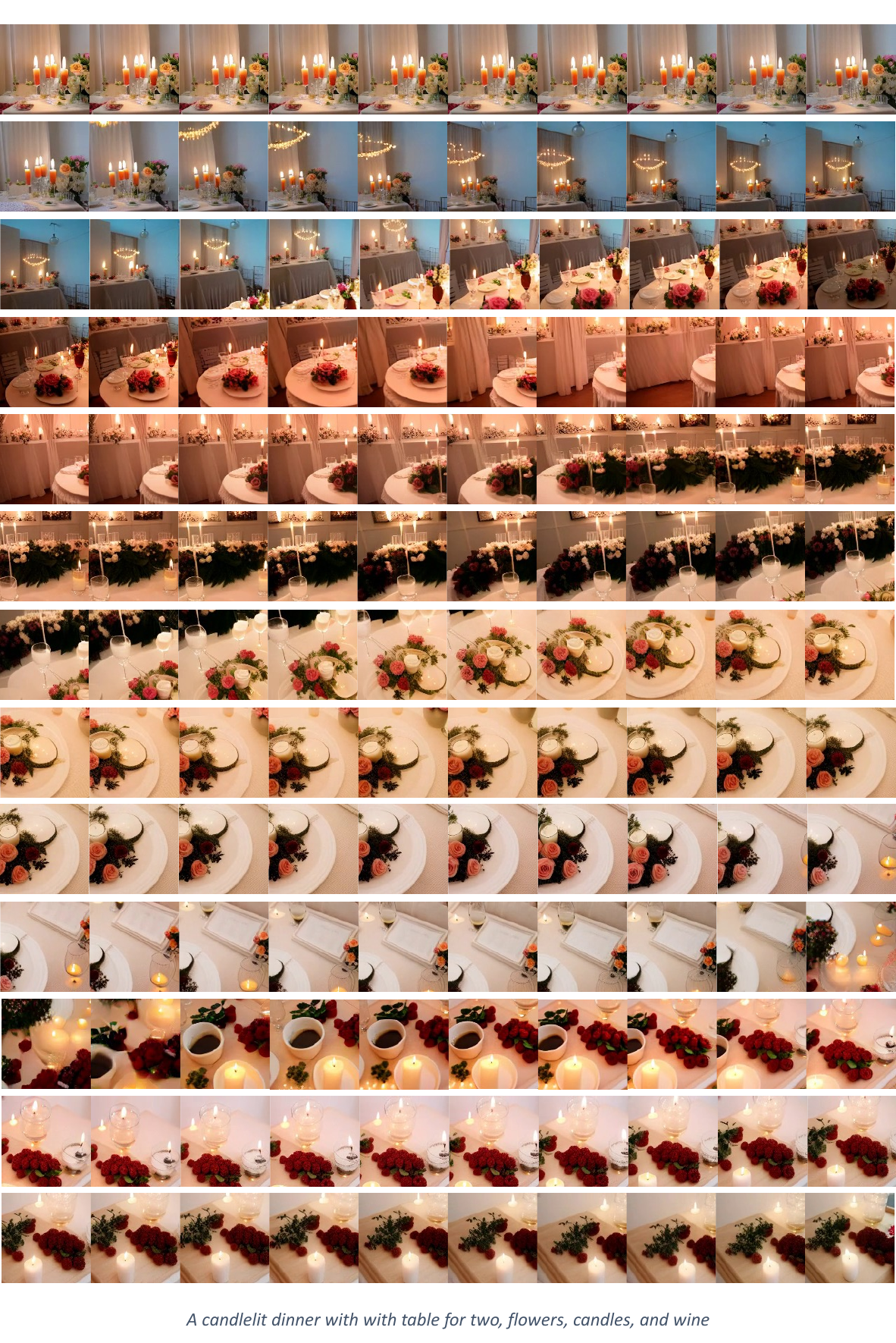}
    \caption{\textbf{Long text-to-video generation results} from \sname. We visualize video frames with a stride of 5. We visualize first 650 frames here and the next 600 frames are visualized in Figure~\ref{fig:t2v_supp_2}.} 
    \label{fig:t2v_supp_1}
\end{figure*}
\begin{figure*}[ht!]
    \centering
    \includegraphics[width=.8\textwidth]{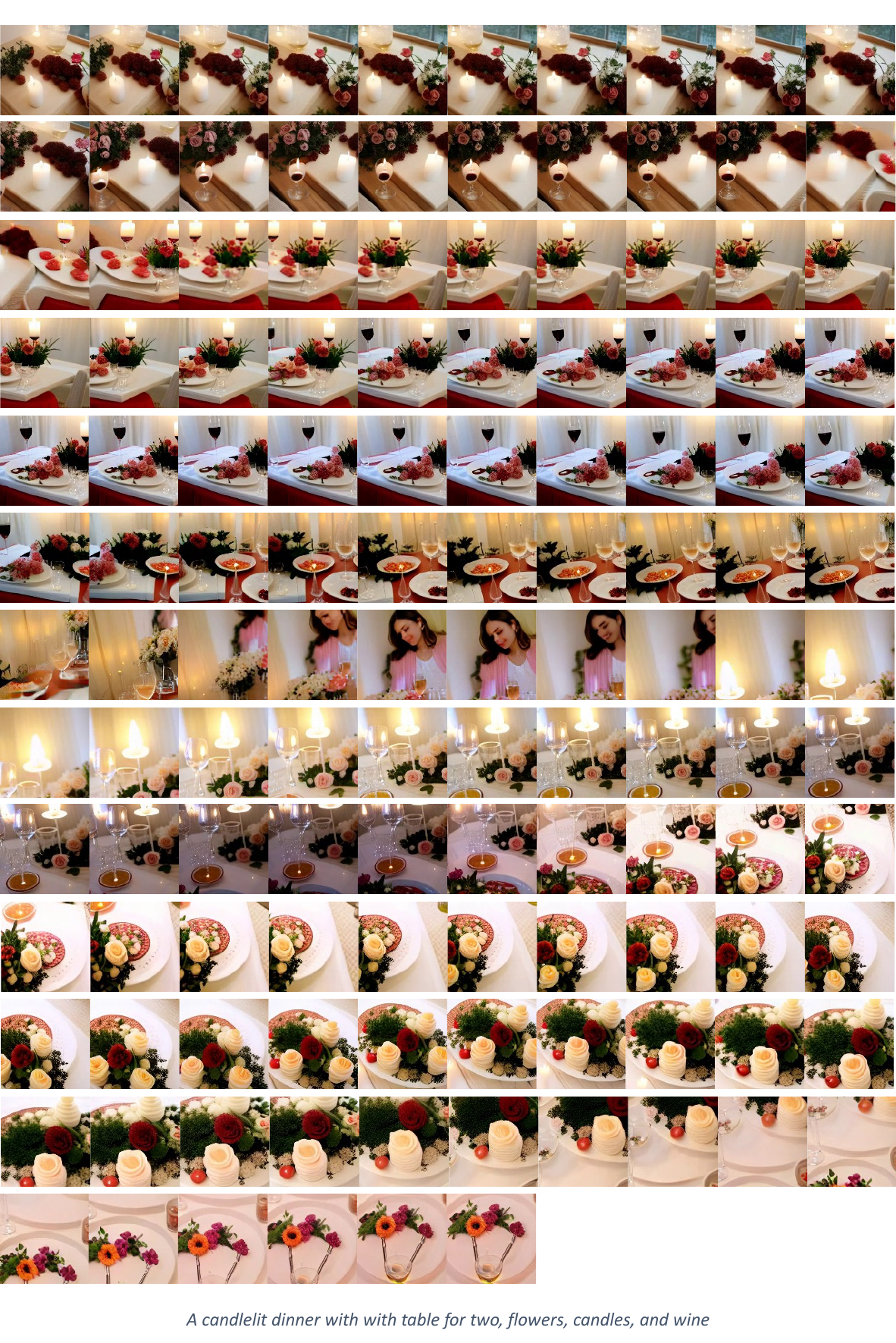}
    \caption{\textbf{Long text-to-video generation results} from \sname. We visualize video frames with a stride of 5. Video frames are continued from Figure~\ref{fig:t2v_supp_1}.
    } 
    \label{fig:t2v_supp_2}
\end{figure*}
\begin{figure*}[ht!]
    \centering
    \includegraphics[width=.8\textwidth]{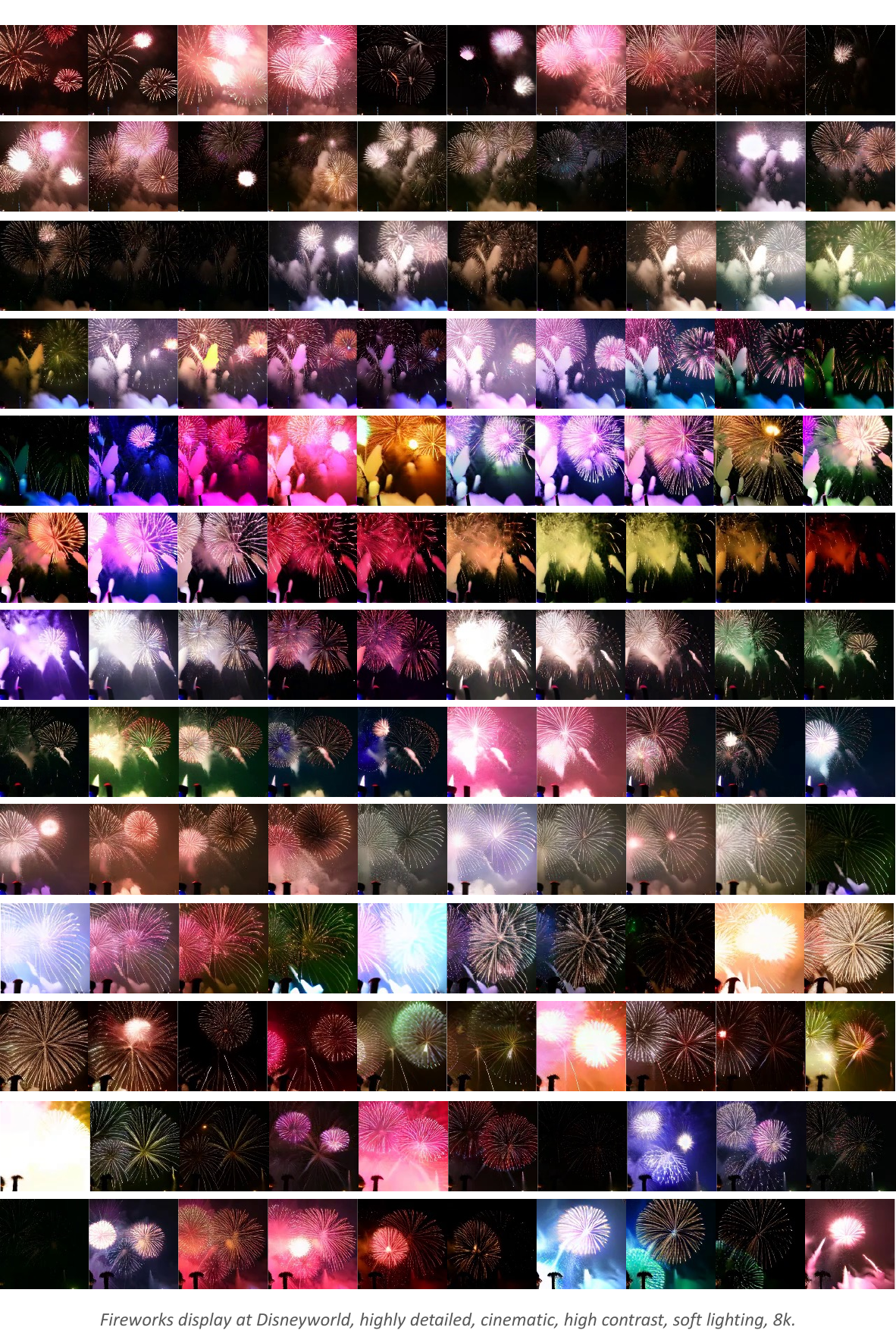}
    \caption{\textbf{Long text-to-video generation results} from \sname. We visualize video frames with a stride of 5. We visualize first 650 frames here and the next 600 frames are visualized in Figure~\ref{fig:t2v_supp_6}.} 
    \label{fig:t2v_supp_5}
\end{figure*}
\begin{figure*}[ht!]
    \centering
    \includegraphics[width=.8\textwidth]{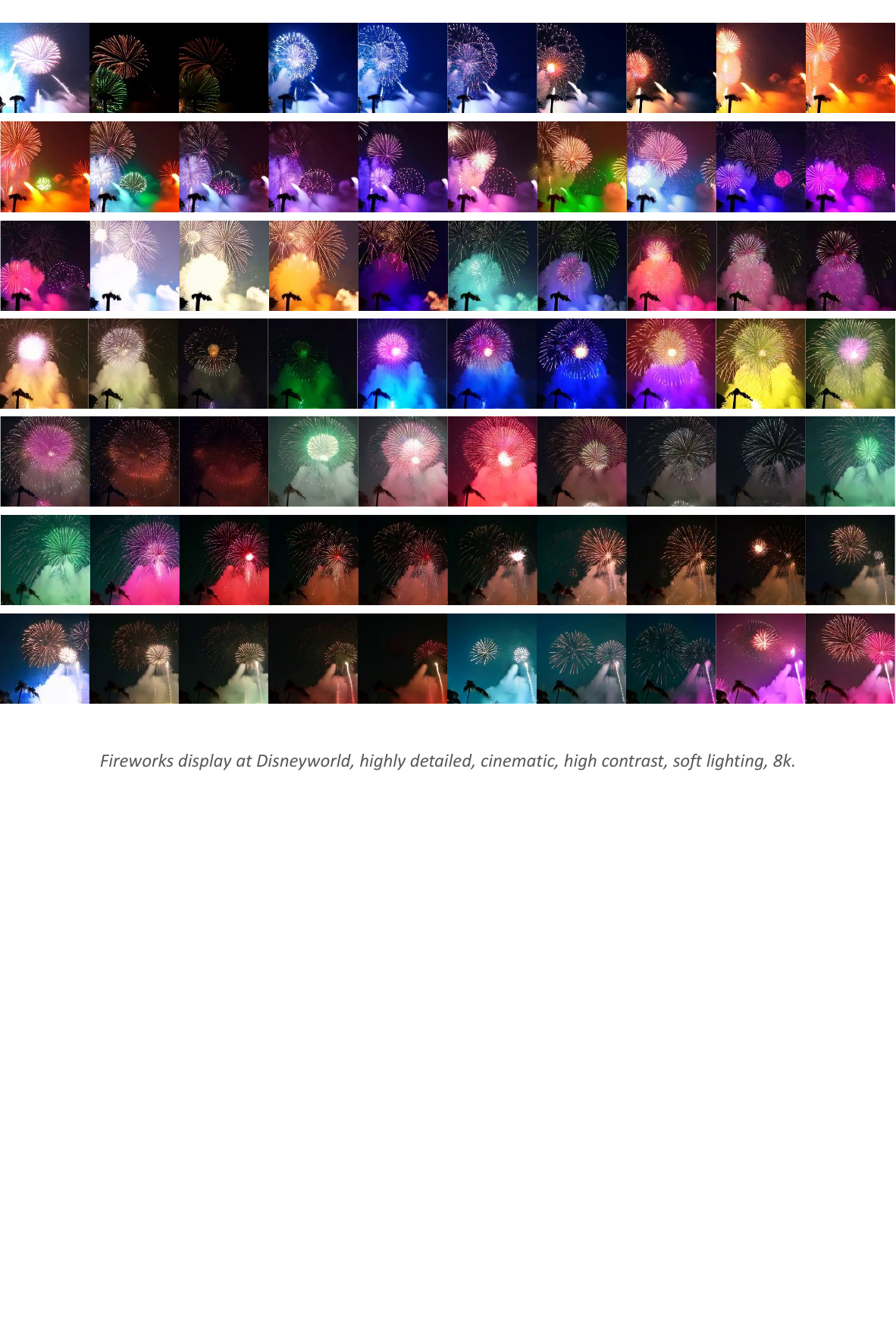}
    \vspace{-3.2in}
    \caption{\textbf{Long text-to-video generation results} from \sname. We visualize video frames with a stride of 5. Video frames are continued from Figure~\ref{fig:t2v_supp_5}.
    } 
    \label{fig:t2v_supp_6}
\end{figure*}
\begin{figure*}[ht!]
    \centering
    \includegraphics[width=.8\textwidth]{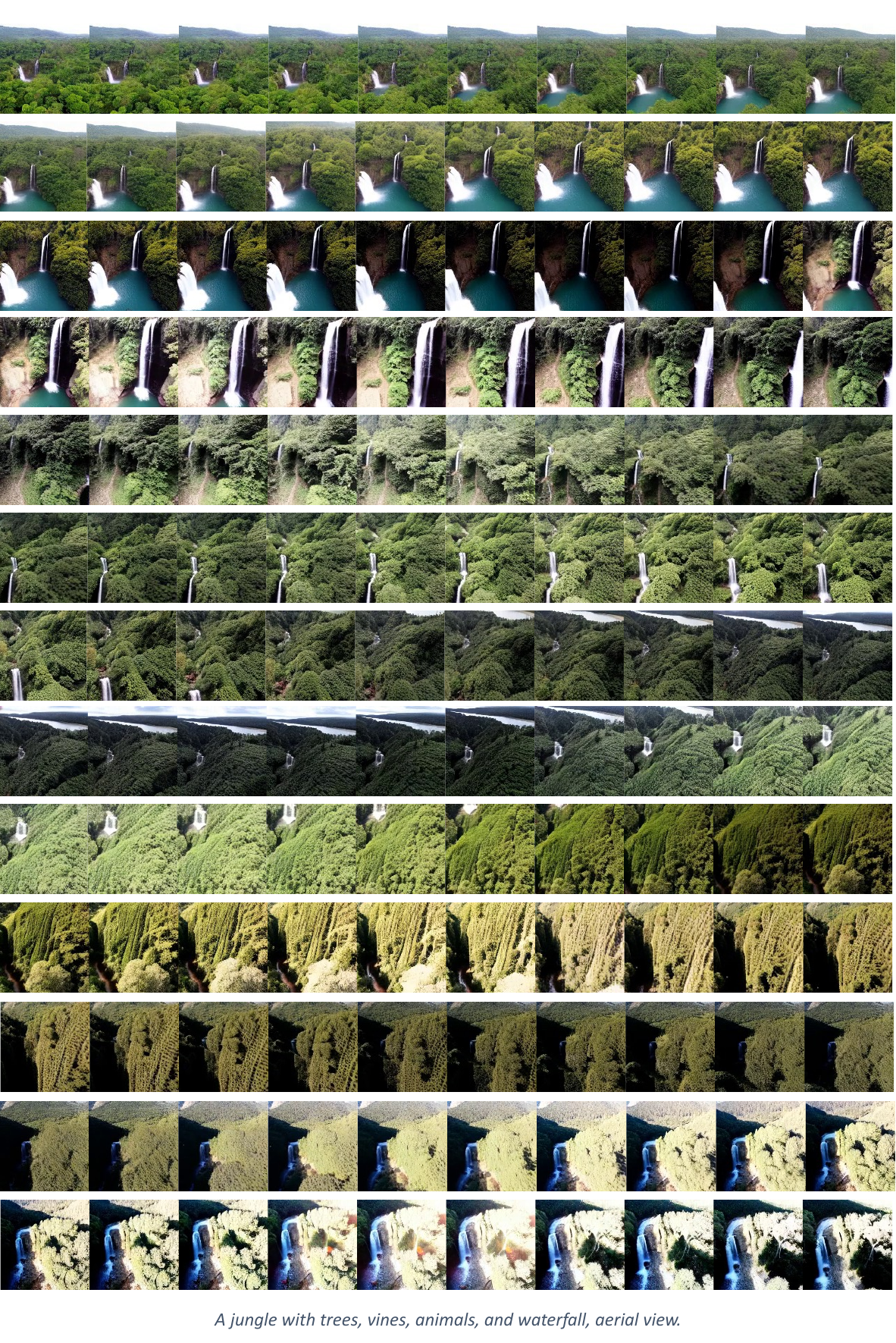}
    \caption{\textbf{Long text-to-video generation results} from \sname. We visualize video frames with a stride of 5. We visualize 650 frames in total.
    } 
    \label{fig:t2v_supp_8}
\end{figure*}
\begin{figure*}[ht!]
    \centering
    \includegraphics[width=.8\textwidth]{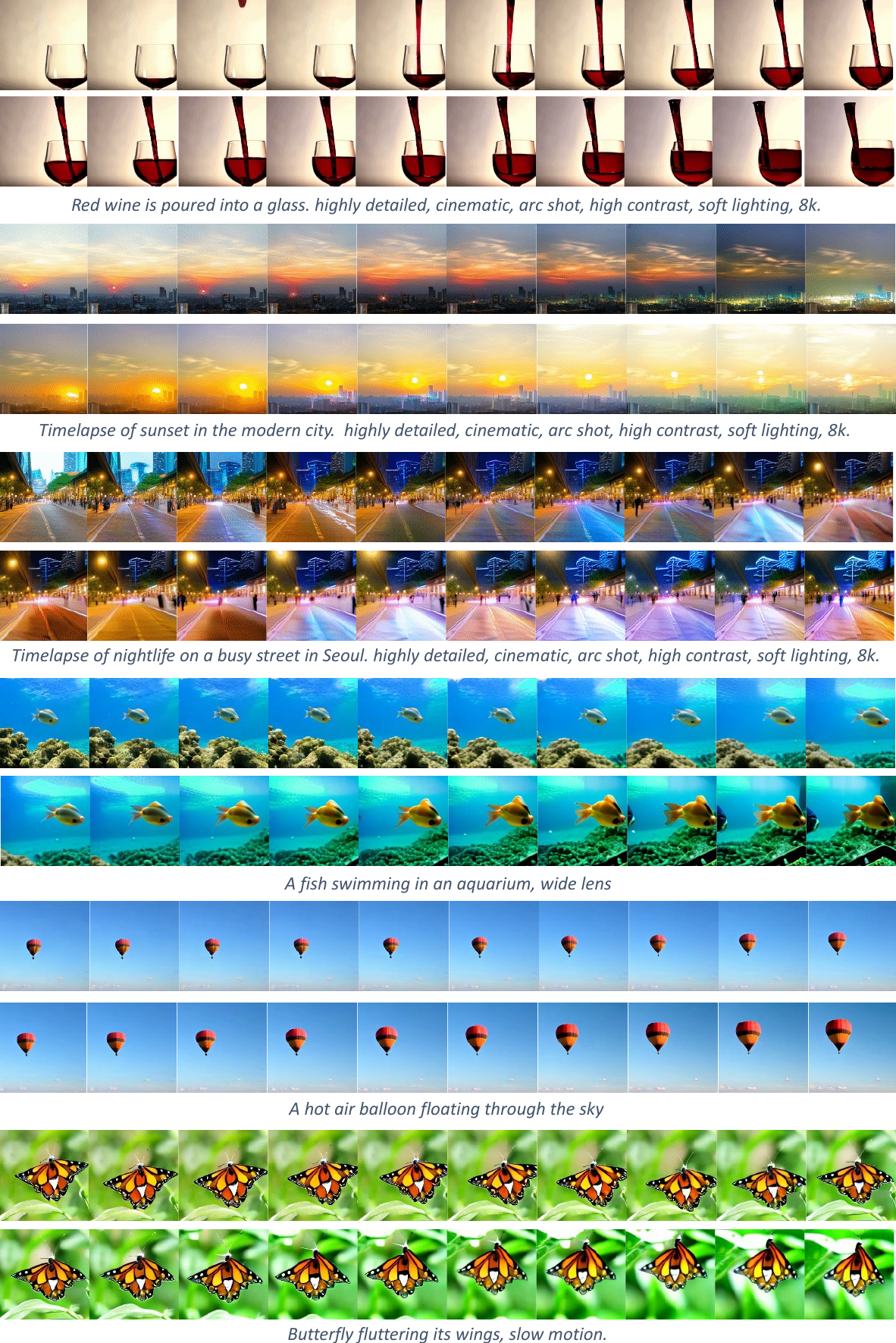}
    \caption{\textbf{Text-to-video generation results} from \sname. We visualize video frames with a stride of 5. We visualize 100 frames in total.
    } 
    \label{fig:t2v_supp_short_1}
\end{figure*}
\begin{figure*}[ht!]
    \centering
    \includegraphics[width=.8\textwidth]{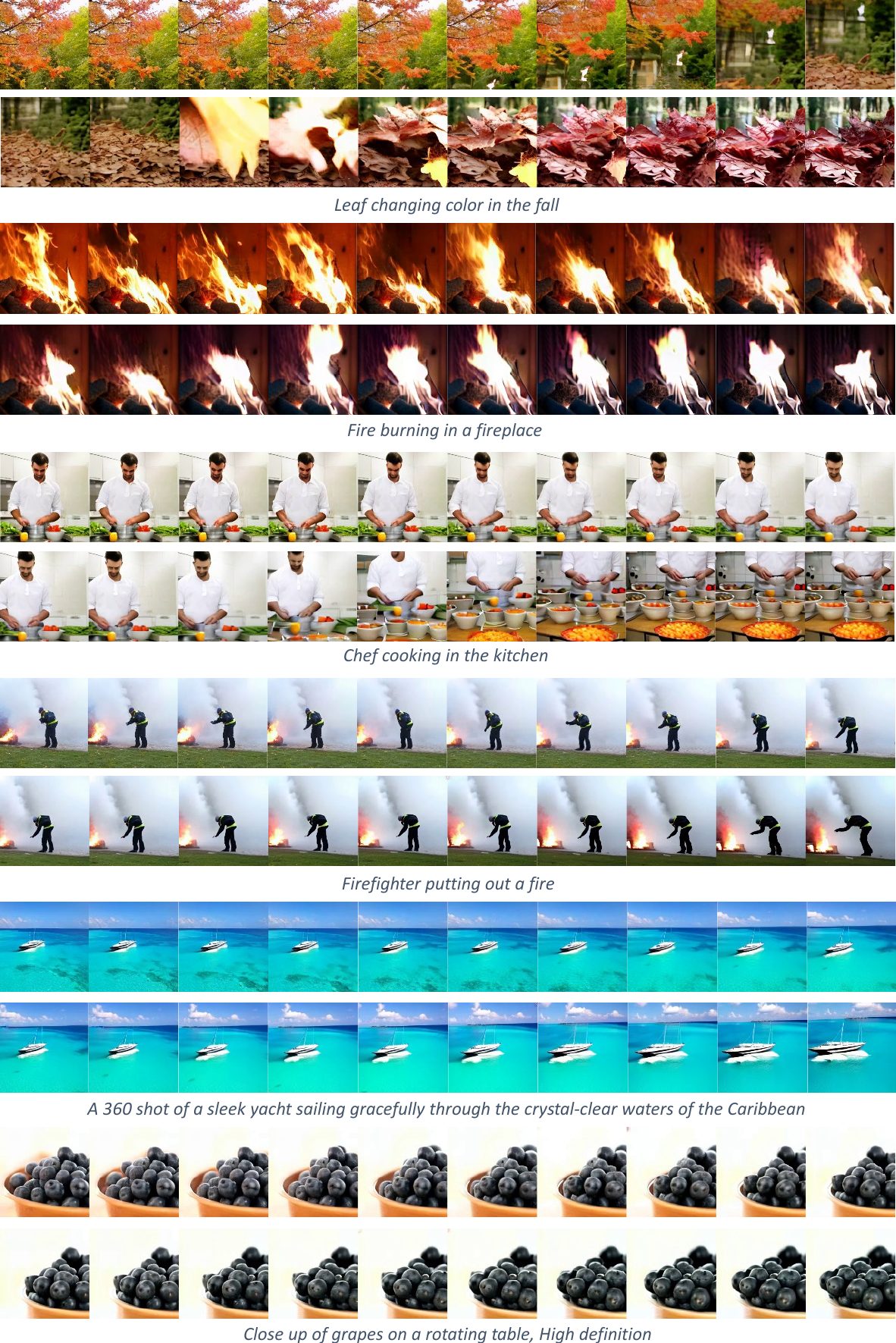}
    \caption{\textbf{Text-to-video generation results} from \sname. We visualize video frames with a stride of 5. We visualize 100 frames in total.
    } 
    \label{fig:t2v_supp_short_2}
\end{figure*}

\clearpage
\clearpage
\clearpage
\clearpage
\section{Limitations and Negative Social Impact}
\label{appen:social_impact}

While \sname shows strong performance in long video generation benchmarks, it is as-yet tested only with a relatively small model.
Considering the scalability of diffusion transformer architectures in text-to-image~\citep{chen2023pixart} or text-to-video generation~\citep{gupta2023photorealistic}, we expect the performance of \sname to show a similar trend given more data and larger models.
Moreover, considering that our text-to-video model is trained on short videos (up to 37 frames), training \sname on large-scale text and long-video paired datasets is a natural next direction.
Finally, one can combine our approach with other techniques for sequence data, such as \citet{ruhe2024rolling}, or consider recent techniques in LLMs to further expand the context window and to leverage its usefulness~\citep{liu2023ring}.

\begin{figure}[ht!]
    \centering    
    \includegraphics[width=.4\textwidth]{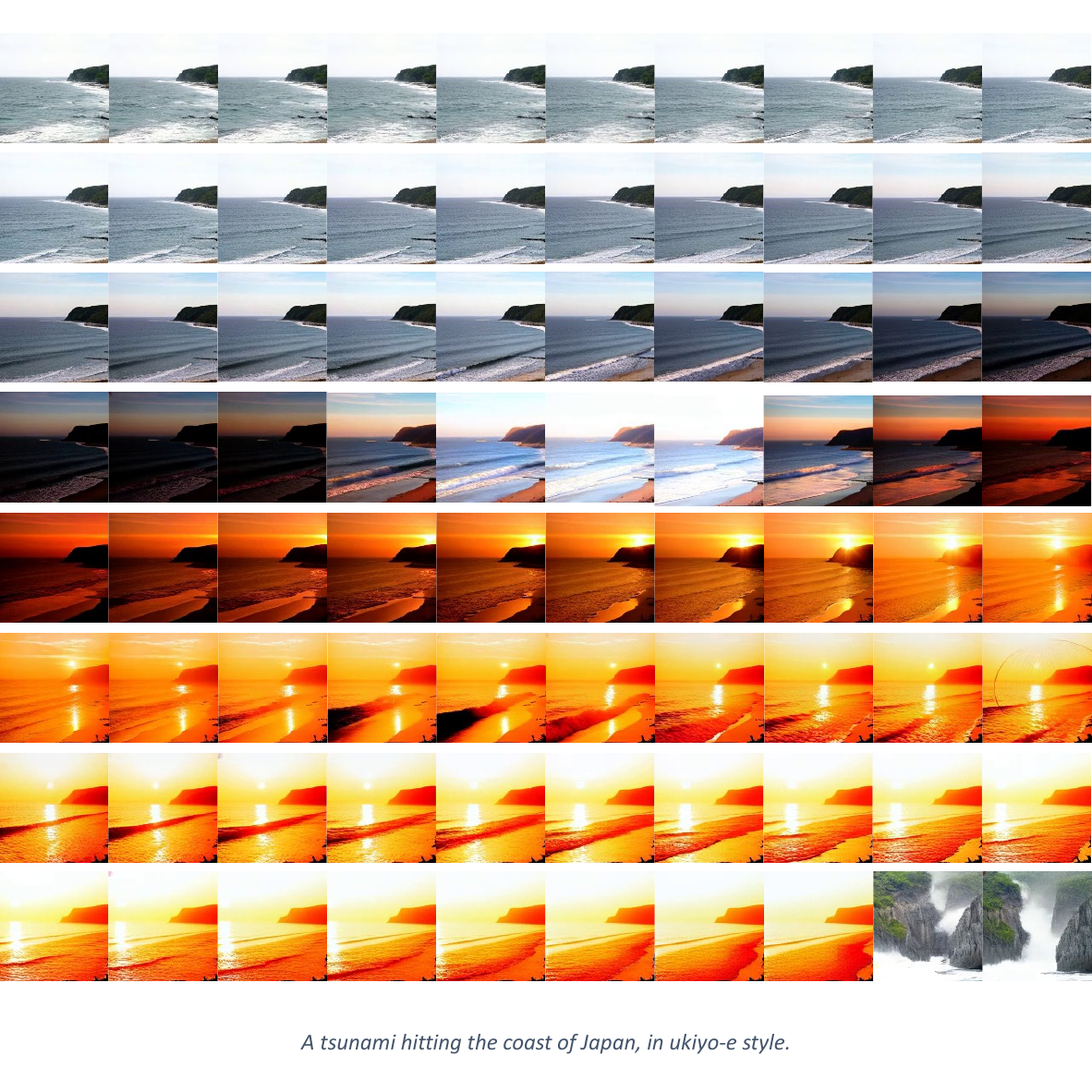}
    \captionof{figure}{\textbf{Failure case} from \sname. We visualize the video frames with a stride of 5. Each video frame has a 128$\times$128 resolution. Text prompts used for generating videos are denoted in the below per each visualization. We visualize first 400 frames.} 
\label{fig:t2v_failure}
\end{figure}
\vspace{0.02in}
\noindent\textbf{Failure cases.}
For text-to-video generation, our model sometimes does not align with the text prompt: for instance, in Figure~\ref{fig:t2v_failure}, the generated video frames do not have ``ukiyo-e'' style, but they are rather photorealistic. Moreover, while \sname shows strong length generalization longer than the training length (\eg, 50 seconds consistent generation in Figure~\ref{fig:main_qual_t2v}), it sometimes fails to generalize and produce uncorrelated frames to the previous context, as shown in the last two frames in Figure~\ref{fig:t2v_failure}. We hypothesize this is because of the small model size and short video length that we used, and we strongly believe training a larger model on long video datasets can solve this issue, considering our superior performance on Kinetics-600 and UCF-101, which are directly trained on long videos.

\vspace{0.02in}
\noindent\textbf{Long-context understanding.}
While \sname shows superior performance in long video generation with efficiency, there might be better diffusion model architecture and formulation to handle long sequences better with large window sizes, and exploring this should be interesting future work. In particular, one may focus on an approach to model extremely long (\eg, 1, 1-hour movies) with diffusion models, which might be suboptimal with \sname because we always assume the fixed-size memory latent vector. One may consider borrowing ideas in recent LLMs into diffusion model contexts, like we have done in this paper, to apply recurrent mechanisms to diffusion models. 


\end{document}